\documentclass[conference]{IEEEtran}
\IEEEoverridecommandlockouts
\usepackage{cite}
\usepackage{amsmath,amssymb,amsfonts}
\usepackage{algorithmic}
\usepackage{graphicx}
\usepackage{textcomp}
\usepackage{xcolor}
\usepackage{makecell}
\usepackage{pifont}
\usepackage[T1]{fontenc}
\usepackage{pdfpages}
\usepackage{subfigure}
\usepackage{multirow}
\usepackage{booktabs}
\usepackage{colortbl}
\usepackage{pifont}
\usepackage{marvosym}
\usepackage{hyperref}
\usepackage{graphicx}
\usepackage{makecell}
\usepackage{multirow}
\usepackage{makecell}
\usepackage{array}     

\def\BibTeX{{\rm B\kern-.05em{\sc i\kern-.025em b}\kern-.08em
    T\kern-.1667em\lower.7ex\hbox{E}\kern-.125emX}}
\begin{document}

\title{CFCH: Coarse-Fine Collaborative Hierarchical Learning for Anterior Segment Disease Analysis\\
}


\author
{\IEEEauthorblockN{Peng Wang}
\IEEEauthorblockA
{\textit{College of Cryptology and Cyber Science} \\
\textit{Nankai University, Tianjin, China}\\
\textit{College of Engineering}  \\
\textit{Yanbian University, Yanji, China}\\
pwang@ybu.edu.cn}
\and

\IEEEauthorblockN{ Haohan Zou}
\IEEEauthorblockA{\textit{Tianjin Eye Hospital} \\
\textit{Nankai University Eye Institute} \\
\textit{Nankai University, Tianjin, China} \\
zouhh1995@163.com}
\and

\IEEEauthorblockN{ Yanlin Wu}
\IEEEauthorblockA
{
\textit{College of
Computer Science } \\
\textit{Nankai University, Tianjin, China}\\
1120230299@nankai.edu.cn}
\and

\IEEEauthorblockN{Xueshuo Xie}
\IEEEauthorblockA{\textit{Haihe Lab of ITAI} \\
Tianjin, China \\
xueshuoxie@nankai.edu.cn}
\and

\IEEEauthorblockN{ Yan Wang\textsuperscript{*}}
\IEEEauthorblockA{\textit{Tianjin Eye Hospital} \\
\textit{Nankai University Eye Institute} \\
\textit{Nankai University, Tianjin, China} \\
wangyan7143@vip.sina.com}
\and

\IEEEauthorblockN{Tao Li\textsuperscript{*}}
\IEEEauthorblockA{\textit{College of Cryptology and Cyber Science} \\
\textit{Nankai University, Tianjin, China}\\
\textit{Haihe Lab of ITAI, Tianjin, China} \\
litao@nankai.edu.cn}
\thanks{\text{*} Corresponding author.}
}

\maketitle

\begin{abstract}
Accurate classification of anterior segment diseases is crucial for ophthalmic screening and diagnosis. 
However, slit-lamp image analysis remains challenging due to substantial variability in imaging conditions and the intrinsic anatomical–disease hierarchy of ocular pathologies. Existing methods typically formulate this task as a flat multi-class classification problem, ignoring the structured dependency between anatomical regions (e.g., cornea, conjunctiva, and lens) and disease manifestations.
To address these limitations, we propose CFCH, a Coarse-Fine Collaborative Hierarchical learning framework that explicitly models anatomical context and disease semantics through a dual-branch architecture. To enable effective cross-granularity collaboration, CFCH introduces semantic and cross-granularity attention consistency constraints, encouraging aligned yet complementary feature learning across branches. In addition, we construct AS-9K, a large-scale anterior segment dataset with 8975 images covering 12 common disease categories. To the best of our knowledge, AS-9K is the largest publicly available dataset for anterior segment image classification.
Extensive experiments on two anterior segment datasets demonstrate that CFCH outperforms state-of-the-art methods. Qualitative visualizations further show more focused and lesion-relevant activation responses, validating the effectiveness of the proposed framework.
Code will be available at https://github.com/ybupengwang/CFCH.
\end{abstract}

\begin{IEEEkeywords}
Anterior Segment, Hierarchical Learning, Consistency Constraints, Medical Image Classification
\end{IEEEkeywords}

\section{Introduction}
\begin{figure}[t]
\centerline
\subfigure{\includegraphics[width=0.485\textwidth]{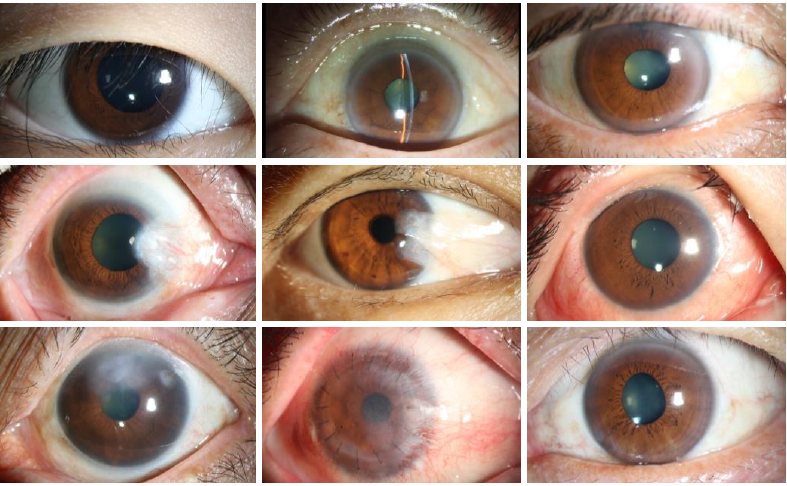}}
\caption{Example slit-lamp images of anterior segment disease categories, showing diverse pathological manifestations across anatomical regions.}
\label{fig:jibing}
\end{figure}

Ophthalmic diseases are generally categorized into anterior segment disorders and fundus diseases \cite{li2021digital}.
Recent advances in deep learning have enabled substantial progress in the automated analysis and computer-assisted diagnosis of fundus diseases \cite{zhou2023foundation,wu2024mm,cai2025retsta,jang2025revisiting,silva2025foundation,zhang2026ai}.
Such progress is largely facilitated by the high level of imaging standardization and structural consistency inherent to fundus imaging modalities \cite{cai2021eyehealer}.
In contrast, the application of artificial intelligence to the automated analysis of anterior segment diseases remains relatively underexplored. Slit-lamp images are acquired using slit-lamp cameras and are routinely used for anterior segment examination, including the eyelids, sclera, conjunctiva, iris, crystalline lens, and cornea \cite{zhang2022machine}. Different anterior segment diseases often require distinct illumination conditions, imaging angles, magnification levels, and observation techniques, leading to pronounced variations in image appearance, spatial structure, and data distribution. As shown in Fig. \ref{fig:jibing}, we present representative lesions across different anatomical structures.

\begin{figure}[t]
\centering
\subfigure{\includegraphics[width=0.5\textwidth]{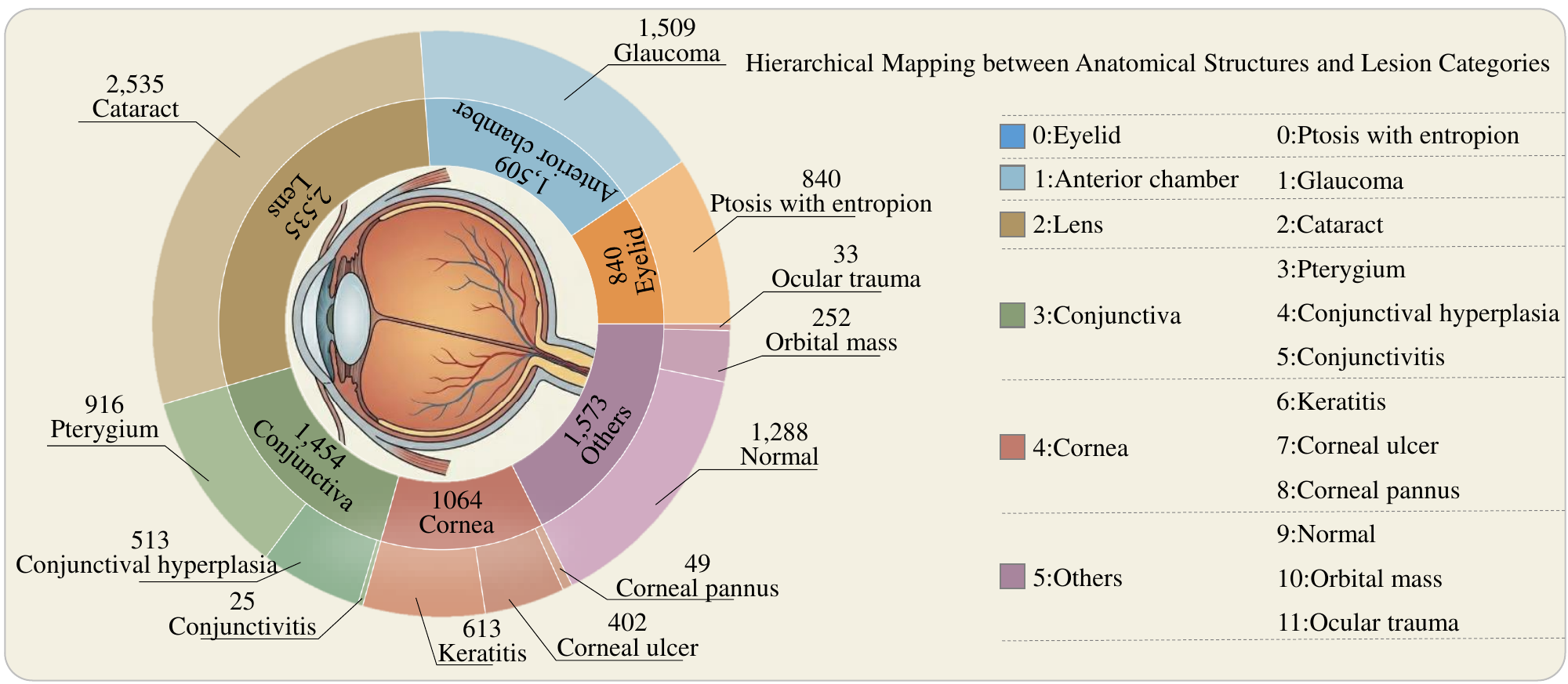}}
\caption{Overview of class distribution and structural–lesion mapping in the AS-9K dataset.}
\label{fig:tongji}
\end{figure}

Consequently, existing diagnostic models are typically designed for disease-specific tasks that involve a single anatomical structure, such as keratitis \cite{kandakji2025hierarchical,li2021preventing}, pterygium \cite{zamania2023pterygium}, conjunctivitis \cite{li2024dual}, or cataract grading \cite{wang2023transformer,xu2019fully}. While effective in constrained settings, such approaches limit clinical applicability. 
When extending to multiple anatomical structures, existing methods often formulate anterior segment disease recognition as a flat multi-class classification problem, where all categories are treated independently without explicitly modeling their underlying anatomical or pathological structure. While these methods have achieved encouraging performance in controlled settings, they ignore the inherent hierarchical organization of anterior segment diseases, where visual patterns are jointly determined by anatomical regions and disease-specific manifestations. To address this limitation, recent studies \cite{ran2023comprehensive,jin2024hmil} have explored hierarchical classification paradigms, which organize visual categories into multiple levels and learn representations in a coarse-to-fine manner. Although these hierarchical approaches can better capture semantic structure compared to flat classification, those methods still independently model different hierarchy levels without sufficient interaction between them. As a result, the coarse-level anatomical representations and fine-level disease-specific features are often learned in isolation, limiting the model’s ability to fully exploit complementary information across granularities and leading to suboptimal feature alignment and representation consistency. In addition, the scarcity of datasets covering diverse common anterior segment diseases further limits the development and evaluation of robust diagnostic models.

\begin{table*}[!t]
    \centering
     \renewcommand\arraystretch{1.0}
    \setlength{\tabcolsep}{5pt}
    \caption{Comparison with existing anterior segment images datasets. \ding{172}, \ding{173}, and \ding{174} represent the detection, segmentation, classification, respectively.}
    \begin{tabular}{lcccccc}
   \toprule
   
        Reference &  Number of Images& Anatomical Structure Coverage& Number of Diseases &   Task & Modality  &  Available  \\ 
        \midrule
    Son et al.(2022) \cite{son2022deep} & 1,355 & Lens & 7 grades & \ding{174}  & slit-lamp & \ding{55}   \\
    Chen et al.(2023) \cite{chen2023automated} & 2,825 &-&3 &   \ding{174}& OCT& \ding{51}   \\
    SLID-E(2024) \cite{dai2024slid}   & 2,999 & -& 4 grades &  \ding{174} &  slit-lamp & \ding{51} \\ 
    Sun et al.(2024) \cite{sun2024oct}   & 1,168 & - &1 &  \ding{173}  & OCT  & \ding{55}  \\
    Ding et al.(2025) \cite{ding2025asdc}   & 1,250 & Eyelid, Conjunctiva, Lens, Cornea  &6 & \ding{174} &  slit-lamp  & \ding{55}  \\       
    SLID(2026) \cite{xu7slid}  & 2,617 & Conjunctiva, Cornea, Pupil&13  & \centering \ding{172} \ding{173} &  slit-lamp   & \ding{51}  \\    
        \rowcolor{gray!20}
        AS-9K(ours) & 8,975 &\makecell{Eyelid, Anterior chamber,  Conjunctiva,\\ Lens, Cornea, Others} &12 &  \ding{174}  & slit-lamp  & \ding{51} \\
        \bottomrule
    \end{tabular}
    \label{tab:dataset}
\end{table*}
To address these issues, we introduce a coarse-fine collaborative hierarchical (CFCH) learning framework for anterior segment image classification. Specifically, CFCH is designed to capture the anatomical constraints and semantic dependencies inherent in ocular pathologies. By integrating a dual-branch architecture for multi-granularity feature learning and a constraint mechanism, our model learns structured and clinically interpretable representations of anterior segment diseases.

\textbf{The main contributions of our work are summarized as follows:} 
\begin{itemize}
 \item We introduce CFCH, a coarse-fine collaborative hierarchical learning framework that jointly models coarse anatomical and fine-grained disease representations for anterior segment image classification.
 \item We propose a HCL strategy that enforces consistency across different granularity levels. By jointly modeling hierarchical semantic consistency and cross-granularity attention alignment.
 \item We construct AS-9K dataset, a large-scale dataset for anterior segment analysis, covering 12 clinically common anterior segment diseases.
\end{itemize}

\section{Dataset}

To support research on automated anterior segment disease analysis, we construct a new large-scale benchmark dataset named AS-9K, specifically designed for slit-lamp image classification.  The dataset aims to facilitate clinically reliable recognition of common anterior segment diseases by providing diverse imaging conditions and expert-verified annotations.
The AS-9K dataset contains 8975 slit-lamp images covering 12 clinically common anterior segment disease categories. Each image is labeled according to the final consensus of experienced ophthalmologists following standard clinical diagnostic guidelines.The dataset is collected from real clinical settings and exhibits a naturally long-tailed distribution, capturing real-world variability in illumination, viewpoint, magnification, and device configuration. As shown in Fig. \ref{fig:tongji}, the class distribution and structural–lesion mapping are presented.

\subsection{Overview}\label{AA}
AS-9K consists of slit-lamp images collected from ophthalmology departments of collaborating hospitals, including both routine screening and clinically confirmed disease cases.
Each sample is annotated with a single disease category label among 12 classes, including keratitis, conjunctivitis, cataract, and other common anterior segment disorders. The label definition follows ophthalmic diagnostic standards and was validated by senior clinicians to ensure diagnostic consistency.
To better reflect real clinical scenarios, no additional preprocessing such as lesion cropping or region annotation was performed. Instead, the dataset preserves the original imaging context, enabling models to learn both global anatomical context and local pathological cues. Importantly, each disease category naturally corresponds to specific anatomical structures, allowing implicit anatomical region inference from disease labels. Therefore, no additional coarse-grained anatomical annotations are required.

The dataset spans a diverse set of pathological categories involving key anterior segment structures, including the cornea, conjunctiva, and crystalline lens, enabling comprehensive evaluation of fine-grained disease classification in anterior segment imaging. As shown in Table \ref{tab:dataset}, compared to existing anterior segment datasets, our dataset addresses several key limitations in both scale and diversity for anterior segment disease recognition. First, AS-9K is the largest slit-lamp dataset, containing 8,975 images, significantly exceeding previous datasets. Second, AS-9K covers 12 clinically common anterior segment diseases and corresponds to 6 anatomical structures, providing broader disease–anatomy coverage than most existing datasets. Third, AS-9K is publicly available, promoting further research in anterior segment disease analysis.


\subsection{Data collection and annotation}
All images were retrospectively collected from slit-lamp examination records between 2019-2024 at participating ophthalmic centers. Patient identifiers were removed prior to dataset construction to ensure privacy protection, and the study was approved by the relevant institutional review boards.

During data screening, ophthalmologists first excluded images with severe blur, overexposure, underexposure, or missing key anatomical regions. However, images with mild artifacts, variations in illumination, or non-standard acquisition angles were intentionally retained to enhance dataset diversity and robustness.
The annotation process was performed in a multi-stage manner. First, each image was independently labeled by two board-certified ophthalmologists. In cases of disagreement, a senior expert with over 10 years of clinical experience performed a final adjudication to determine the ground-truth label. This consensus-based strategy ensures high annotation reliability while reducing inter-observer variability.

\begin{figure*}[!t]
\centering
\subfigure{\includegraphics[width=0.8\textwidth]{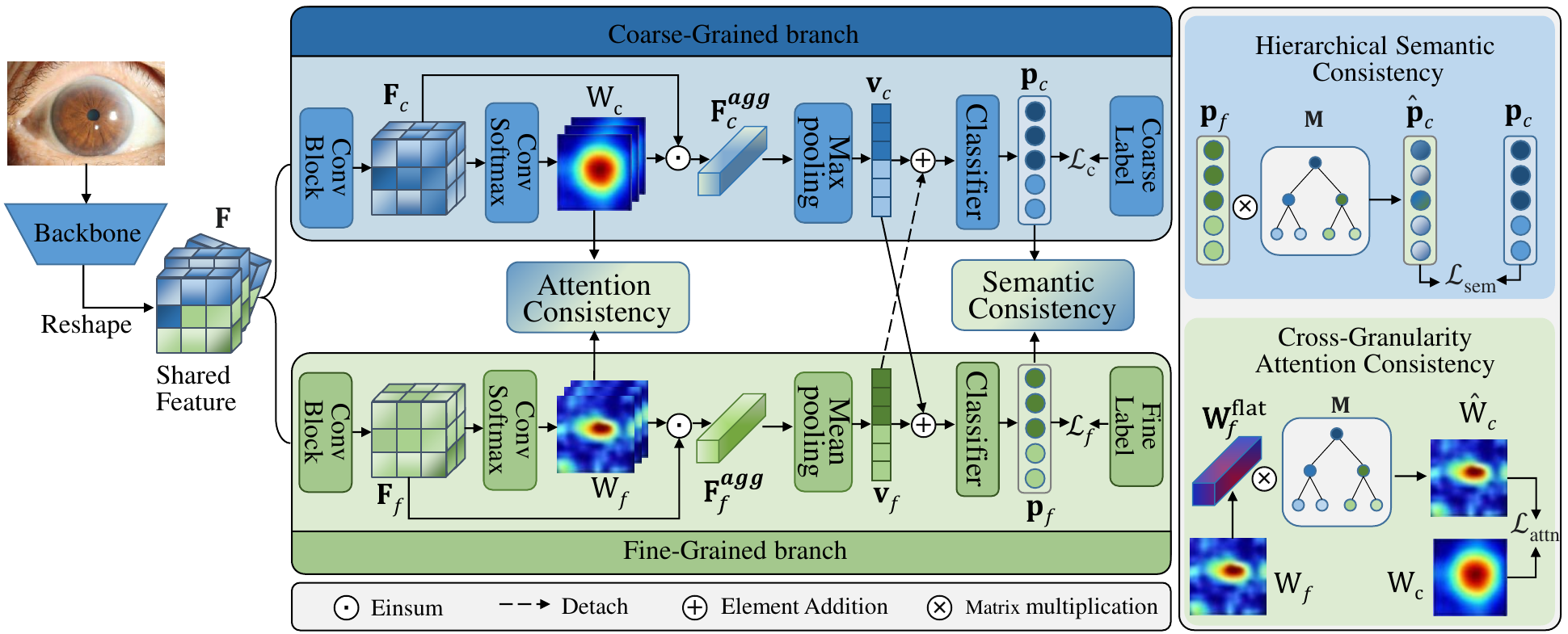}}
\caption{Overall structure of the proposed CFCH, showing the shared feature extractor, multi-granularity learning module, and hierarchical constraint mechanism. Coarse-grained labels correspond to anatomical structure categories, while fine-grained labels represent disease categories.}
\label{fig:overall}
\end{figure*}

\section{Method}
The proposed framework is illustrated in  Fig. \ref{fig:overall}. It consists of three main components: (1) a shared feature extractor based on ViT \cite{dosovitskiy2020image}, (2) a multi-granularity learning module with dual branches for coarse-grained anatomical classification branch and a fine-grained disease classification branch, and (3) a hierarchical constraint mechanism that enforces consistency across different granularity levels through both semantic and spatial attention alignment.

\subsection{Multi-Granularity Feature Learning}
To capture discriminative features at different granularity levels, we design two parallel convolutional branches that process the shared features $\mathbf{F}$ independently:

\begin{equation}
\mathbf{F}_{\{c,f\}} = \phi_{3\times3} \left( 
\sigma \left( 
\mathrm{BN} \left( \phi_{1\times1} (\mathbf{F}) \right) 
\right) 
\right)
\end{equation}
where $\mathbf{F} \in \mathbb{R}^{C \times h \times w}$ is the backbone feature map, with $C$ channels and spatial size $h \times w$. $\mathbf{F}_c$ and $\mathbf{F}_f $ represent the feature representations at the coarse and fine levels, respectively.
$\phi_{1\times1}$ and $\phi_{3\times3}$ represent the $1\times1$ and $3\times3$ convolutional operations, $\text{BN}$ is Batch Normalization, and $\sigma$ denotes the ReLU activation function.
The parallel design allows each branch to learn granularity-specific representations while sharing the same backbone features, enabling effective knowledge transfer between different levels.

Inspired by the observation that different diseases manifest in distinct spatial regions (e.g., pterygium typically appears at the corneal limbus), we introduce a class-specific spatial attention mechanism to highlight diagnostically relevant regions for each category.

\begin{equation}
\mathbf{W}_{\{c,f\}} =
\mathrm{Softmax} \left(
\phi_{1\times1} \left( \mathbf{F}_{\{c,f\}} \right)
\right)
\end{equation}
where $\mathbf{W}_{\{c,f\}} \in \mathbb{R}^{N_{\{c,f\}} \times h \times w}$ represents
the class-specific spatial attention maps at the coarse and fine levels,
respectively, where $N_c$ and $N_f$ denote the numbers of coarse-grained and fine-grained categories.

To adaptively aggregate class-specific features while preserving semantic consistency across different granularity levels,  we employ an attention-guided spatial aggregation. Specifically, discriminative features are aggregated by performing a weighted summation of the spatial descriptors $\mathbf{F}_{\{c,f\}}$ guided by
their corresponding class-specific salience maps $\mathbf{W}_{\{c,f\}}$:

\begin{equation} 
\mathbf{F}_{\{c,f\}}^{\mathrm{agg}} = \sum_{h=1}^H \sum_{w=1}^W \mathbf{W}_{\{c,f\}}^{(n,h,w)} \cdot \mathbf{F}_{\{c,f\}}^{(d,h,w)} \end{equation}
where $n \in \{1, \dots, N_{\{c,f\}}\}$ and $d \in \{1, \dots, C\}$ denote the category index and feature channel dimension, respectively. The resulting aggregated feature $\mathbf{F}_c^{\mathrm{agg}} \in \mathbb{R}^{N_c \times C}$ and $\mathbf{F}_f^{\mathrm{agg}} \in \mathbb{R}^{N_f \times C}$ capture discriminative semantic information for anatomical categories and fine-grained pathological classes.

To obtain representative feature vectors for each hierarchical branch, we employ distinct pooling strategies along the class dimension.
Max pooling is applied to the coarse-grained branch to highlight the most discriminative anatomical responses, whereas mean pooling is used in the fine-grained branch to aggregate complementary pathological features.
The resulting hierarchy-level feature vectors are thus defined as:

\begin{equation}
\mathbf{v}_c = \max_{n=1,\dots,N_c} \mathbf{F}_{c,n}^{\mathrm{agg}}, \quad
\mathbf{v}_f = \frac{1}{N_f} \sum_{n=1}^{N_f} \mathbf{F}_{f,n}^{\mathrm{agg}}
\end{equation}

where $\mathbf{v}_c \in \mathbb{R}^{C}$ and $\mathbf{v}_f \in \mathbb{R}^{C}$ denote the output features for the coarse-grained and fine-grained branches, respectively. 
To facilitate information exchange between granularity levels, we use asymmetric feature fusion that blocks coarse classification gradients to fine-grained features through the fusion path, while allowing fine classification gradients to update both representations.

\begin{equation} 
\mathbf{y}_{c} = \text{Classifier}_{c}(\text{ReLU}(\mathbf{v}_{c} + \text{detach}(\mathbf{v}_{f})))
\end{equation}
\begin{equation} 
\mathbf{y}_{f} = \text{Classifier}_{f}(\text{ReLU}(\mathbf{v}_{f} + \mathbf{v}_{c}))
\end{equation}

\subsection{Hierarchical Constraint Learning}
To explicitly encode the semantic hierarchy between anatomical regions and fine-grained disease categories, we introduce a hierarchical constraint learning (HCL) strategy.
HCL enforces consistency across prediction semantics and spatial evidence by jointly modeling hierarchical semantic consistency and cross-granularity attention consistency.

\noindent \textbf{Hierarchical Semantic Consistency.}
Although the coarse-grained and fine-grained branches are supervised independently, unconstrained optimization may lead to semantically inconsistent predictions across hierarchical levels.
To address this issue, we impose a hierarchical semantic consistency constraint that explicitly aligns predictions across granularities.
Let $\mathbf{p}_{c}$ and $\mathbf{p}_{f} $ denote the Softmax probability vectors of $\mathbf{y}_{c}$ and $\mathbf{y}_{f}$, respectively.
We define a hierarchical mapping matrix $\mathbf{M} \in \mathbb{R}^{N_c \times N_f}$, where $\mathbf{M}_{ij}=1$ if fine-grained class $j$ belongs to coarse-grained class $i$, and $0$ otherwise.

The fine-grained probabilities are aggregated to the coarse level as:
\begin{equation}
\hat{\mathbf{p}_c}
=
\mathbf{p}_f \mathbf{M}^{\top}
\end{equation}

We then enforce semantic alignment between the projected coarse probabilities and the coarse branch predictions by minimizing:
\begin{equation}
\mathcal{L}_{\mathrm{sem}}
=
\left\|
\hat{\mathbf{p}_c}
-
\mathbf{p}_c
\right\|_2^2
\end{equation}
This constraint encourages semantic coherence across hierarchies, ensuring that the predicted probability of an anatomical region is consistent with the collective confidence of its associated fine-grained diseases.

\noindent \textbf{Cross-Granularity Attention Consistency.}
Correct hierarchical predictions should also be supported by anatomically plausible spatial evidence.
To this end, we introduce a cross-granularity attention consistency constraint that aligns attention distributions between the fine-grained and coarse-grained branches.
We reshape the fine-grained attention maps $\mathbf{W}_f \in \mathbb{R}^{N_f\times h \times w}$ into $\mathbf{W}_{f}^{flat} \in \mathbb{R}^{N_f \times (h \cdot w)}$ and project them to the coarse level using the $\mathbf{M}$:
\begin{equation}
\hat{\mathbf{W}_{c}}
=
\mathbf{M}\,\mathbf{W}_{f}^{\mathrm{flat}}
\end{equation}
The cross-granularity attention consistency loss is defined as:
\begin{equation}
\mathcal{L}_{\mathrm{attn}}
=
\left\|
\hat{\mathbf{W}_{\mathrm{c}}}
-
\mathbf{W}_{\mathrm{c}}
\right\|_2^2
\end{equation}

By enforcing this constraint, fine-grained disease-specific attention is spatially bounded within the corresponding anatomical regions highlighted by the coarse branch.
This effectively prevents the model from making correct predictions based on anatomically irrelevant or spurious regions, thereby enhancing interpretability and robustness.

The final optimization objective integrates hierarchical classification supervision and both consistency constraints:
\begin{equation}
\mathcal{L}_{\mathrm{total}}
=
\mathcal{L}_{f}
+
\lambda_{c}  \mathcal{L}_{c}
+
\lambda_{\mathrm{s}} \mathcal{L}_{\mathrm{sem}}
+
\lambda_{\mathrm{a}} \mathcal{L}_{\mathrm{attn}}
\end{equation}
where $\lambda_{c}$, $\lambda_{s}$ and $\lambda_{a}$ balance the contributions of hierarchical semantic consistency and cross-granularity attention consistency, respectively.

\section{Experiment}
\subsection{Dataset and Evaluation Metrics}
\noindent \textbf{SLID Dataset} 
SLID \cite{xu7slid} collected and released a slit-lamp anterior segment image dataset consisting of 2,617 clinical photographs. The dataset covers a wide spectrum of anterior segment conditions, including 245 normal images, 2,091 monomorbidity images, and 281 multimorbidity images. In this study, we focus exclusively on the 2,091 monomorbidity images. The dataset is split at the patient level, with 80\% used for training and the remaining 20\% for testing.

\noindent \textbf{AS-9K Dataset}
We collected a dedicated anterior segment image dataset comprising 8,975 slit-lamp images, covering 12 common anterior segment diseases. All images were acquired under routine clinical settings using slit-lamp cameras, ensuring realistic variations in illumination, viewpoint, and disease manifestation. The annotated dataset was split into training and testing subsets with an 8:2 ratio using stratified sampling based on disease categories.

\noindent \textbf{Evaluation Metrics} We evaluate performance using the metrics: Accuracy (Acc), Precision, Recall, and F1-score.
\begin{figure*}[!t]
\centering
\setlength{\tabcolsep}{2pt}
\resizebox{\textwidth}{!}{
\begin{tabular}{ccccc}
 ResNet & ViT &  HiFuse & ADSR & Ours \\
\includegraphics[width=0.19\textwidth, trim=48 35 70 20, clip]{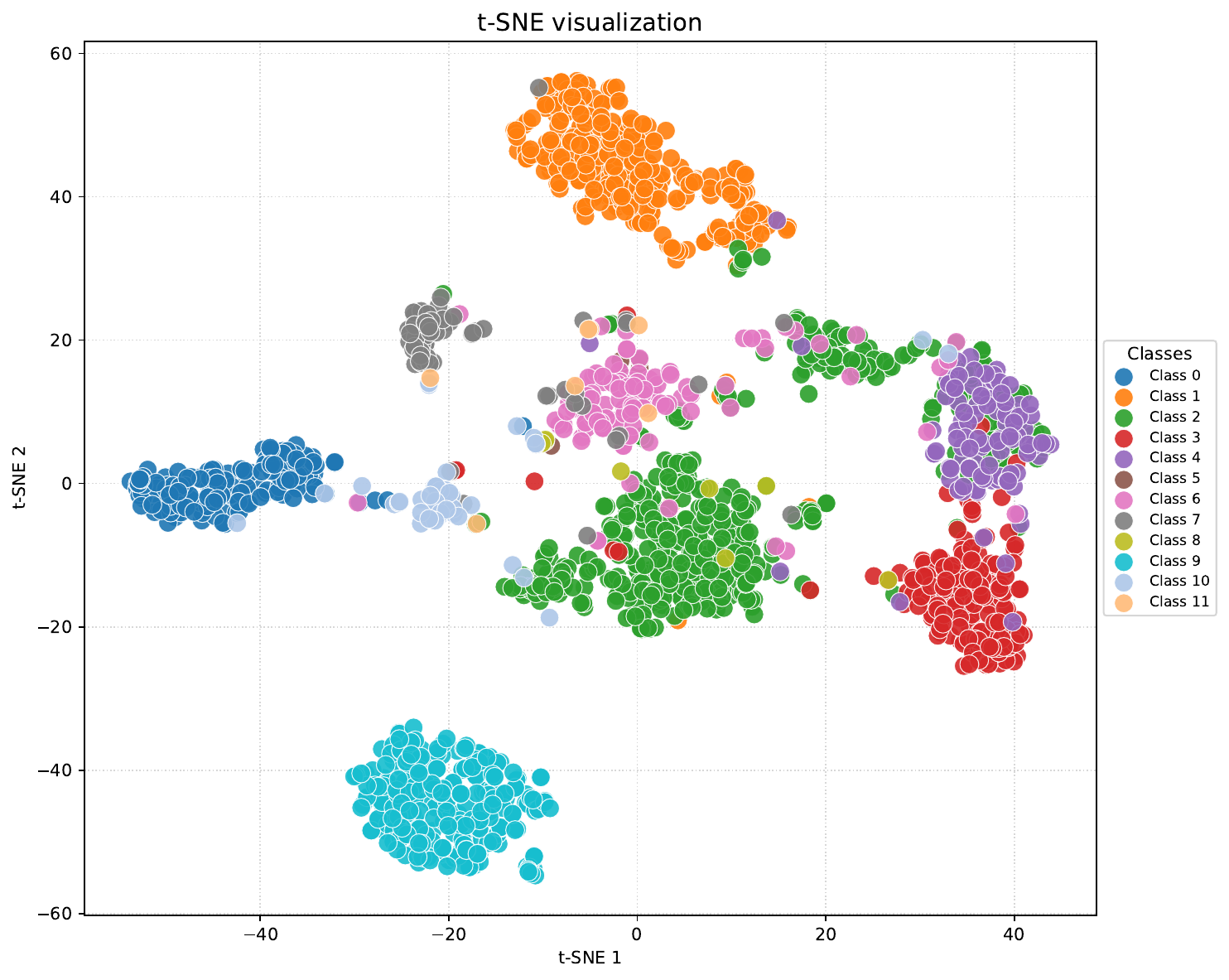} &
\includegraphics[width=0.19\textwidth, trim=48 35 70 20, clip]{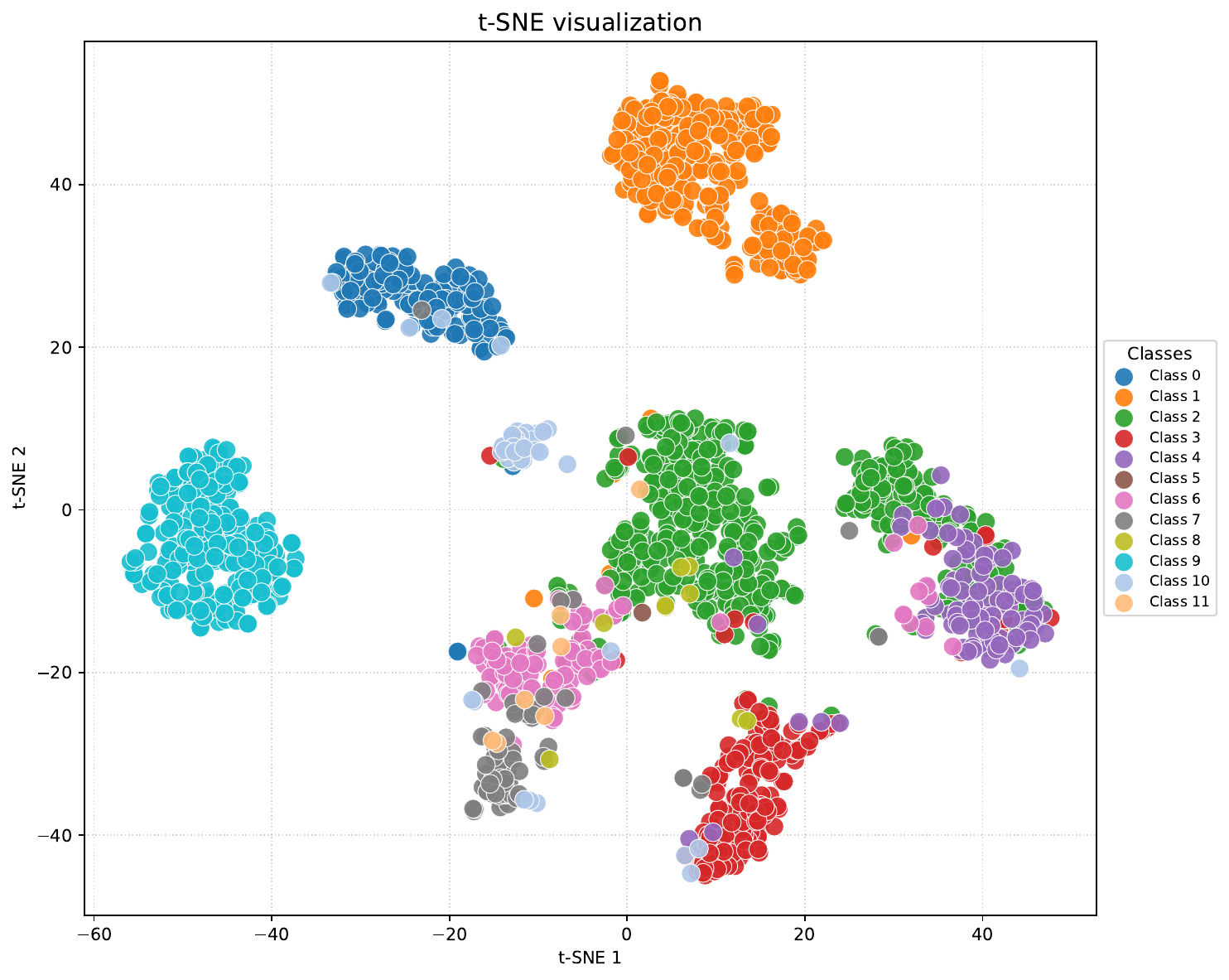} &
\includegraphics[width=0.19\textwidth, trim=48 35 70 20, clip]{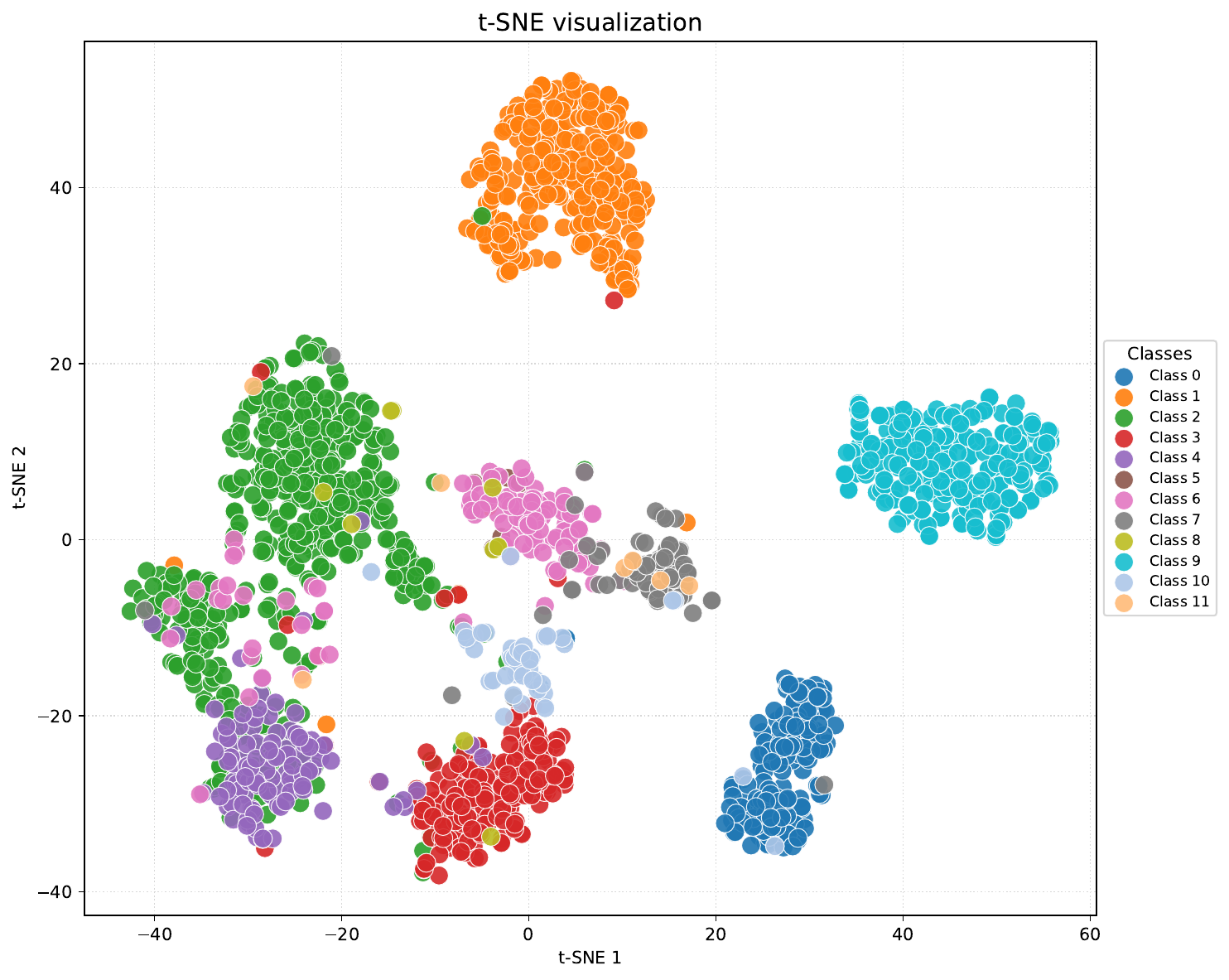} &
\includegraphics[width=0.19\textwidth, trim=48 35 70 20, clip]{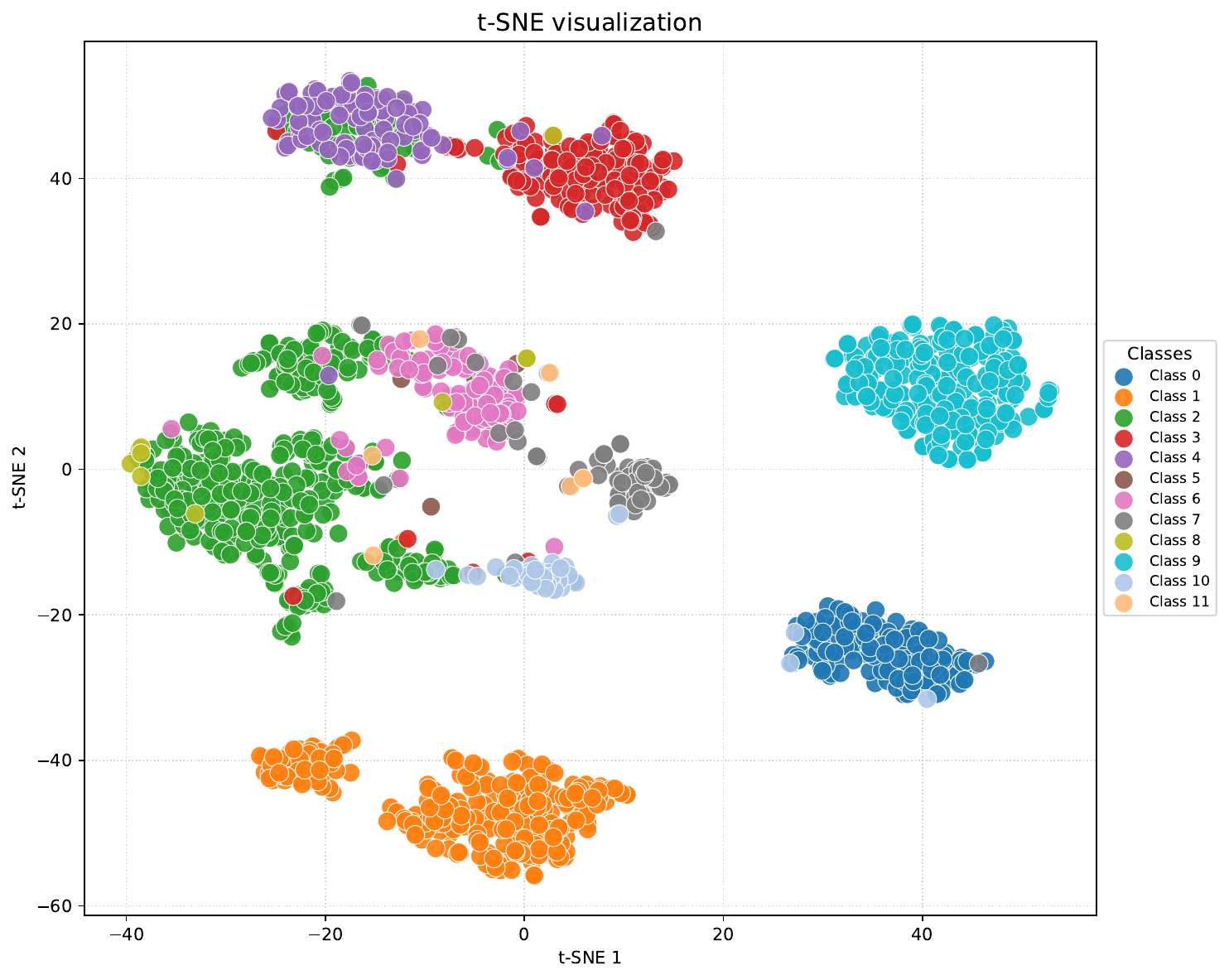} &
\includegraphics[width=0.19\textwidth, trim=48 35 70 20, clip]{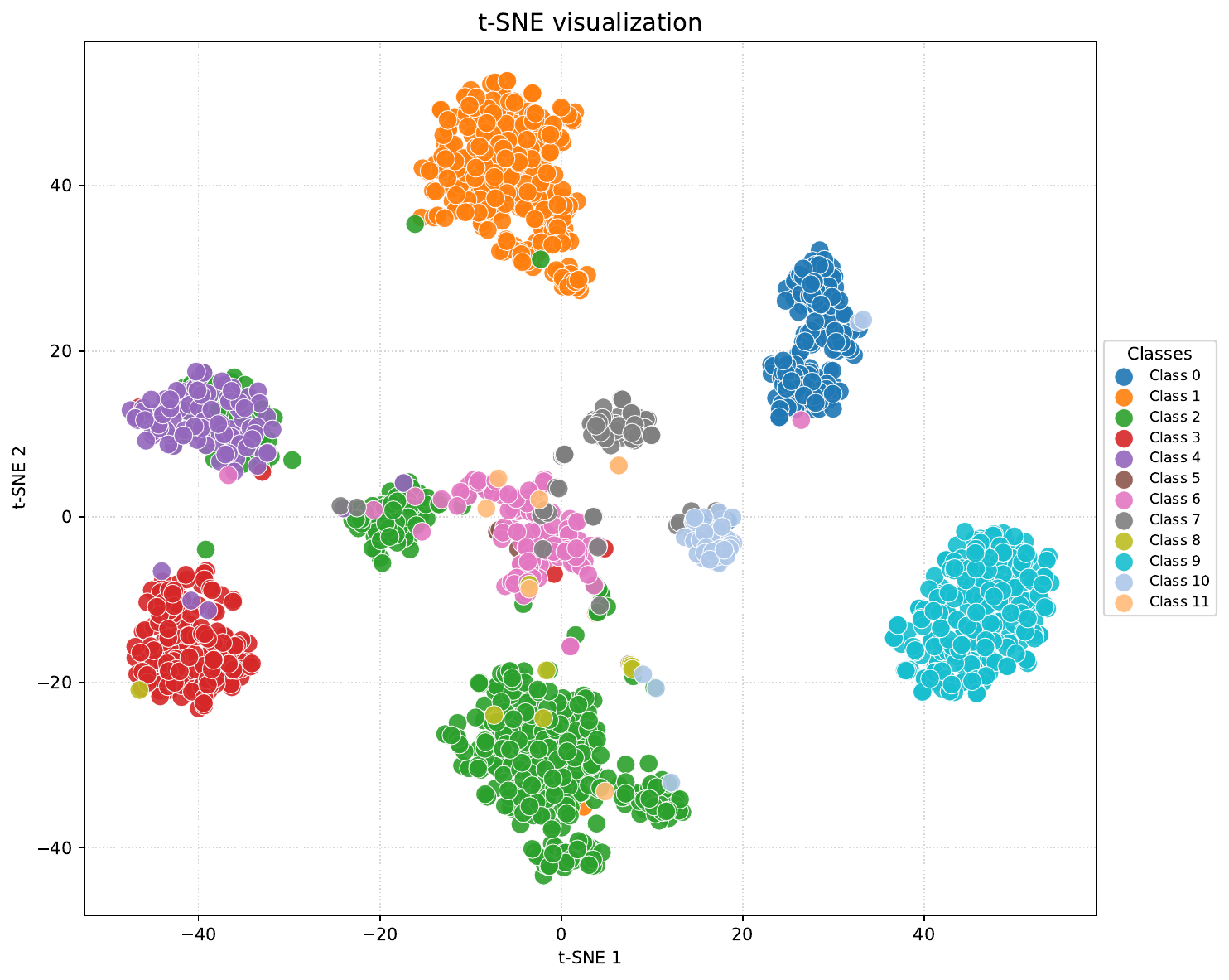} \\

\includegraphics[width=0.19\textwidth, trim=48 35 70 20, clip]{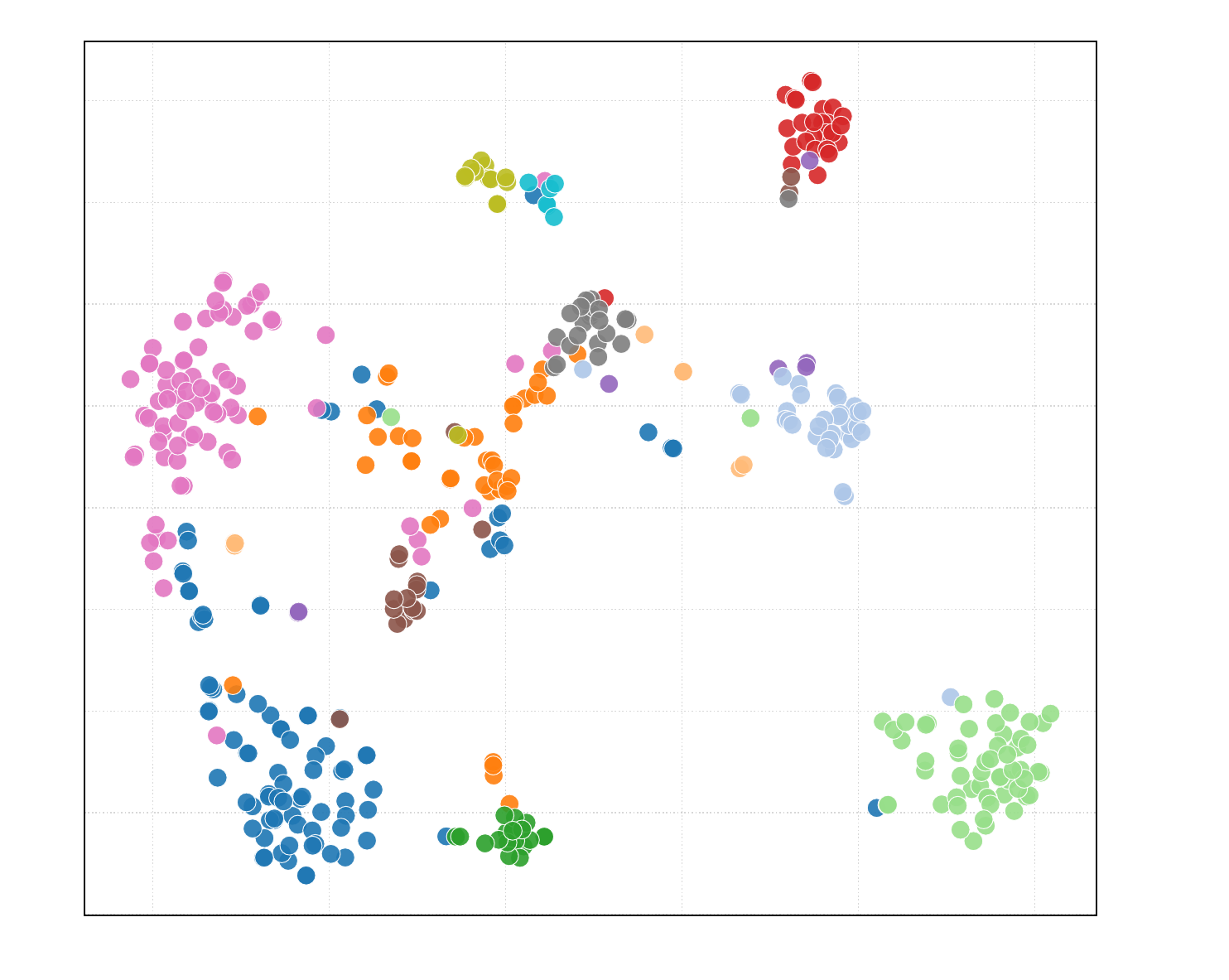} &
\includegraphics[width=0.19\textwidth, trim=48 35 70 20, clip]{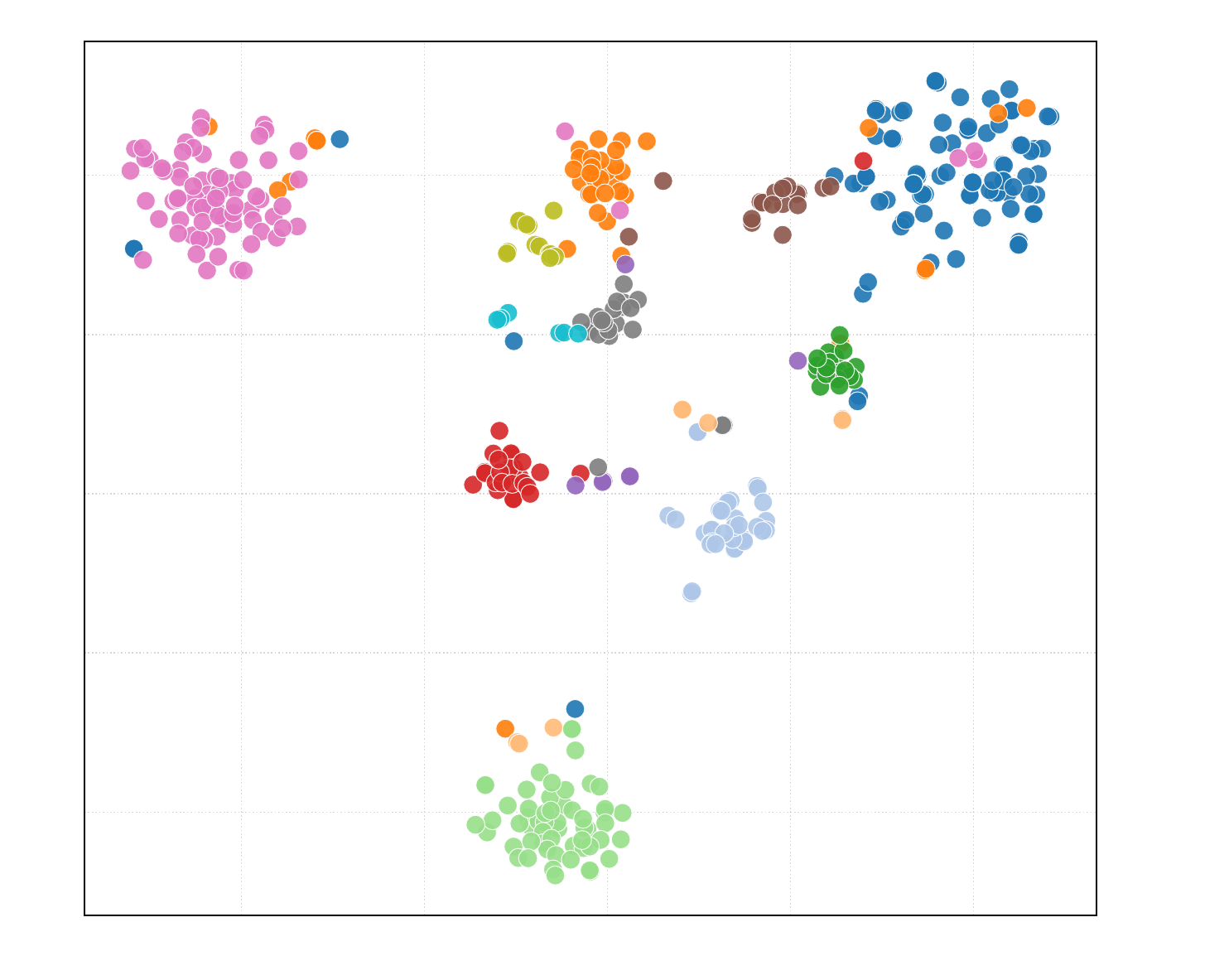} &
\includegraphics[width=0.19\textwidth, trim=48 35 70 20, clip]{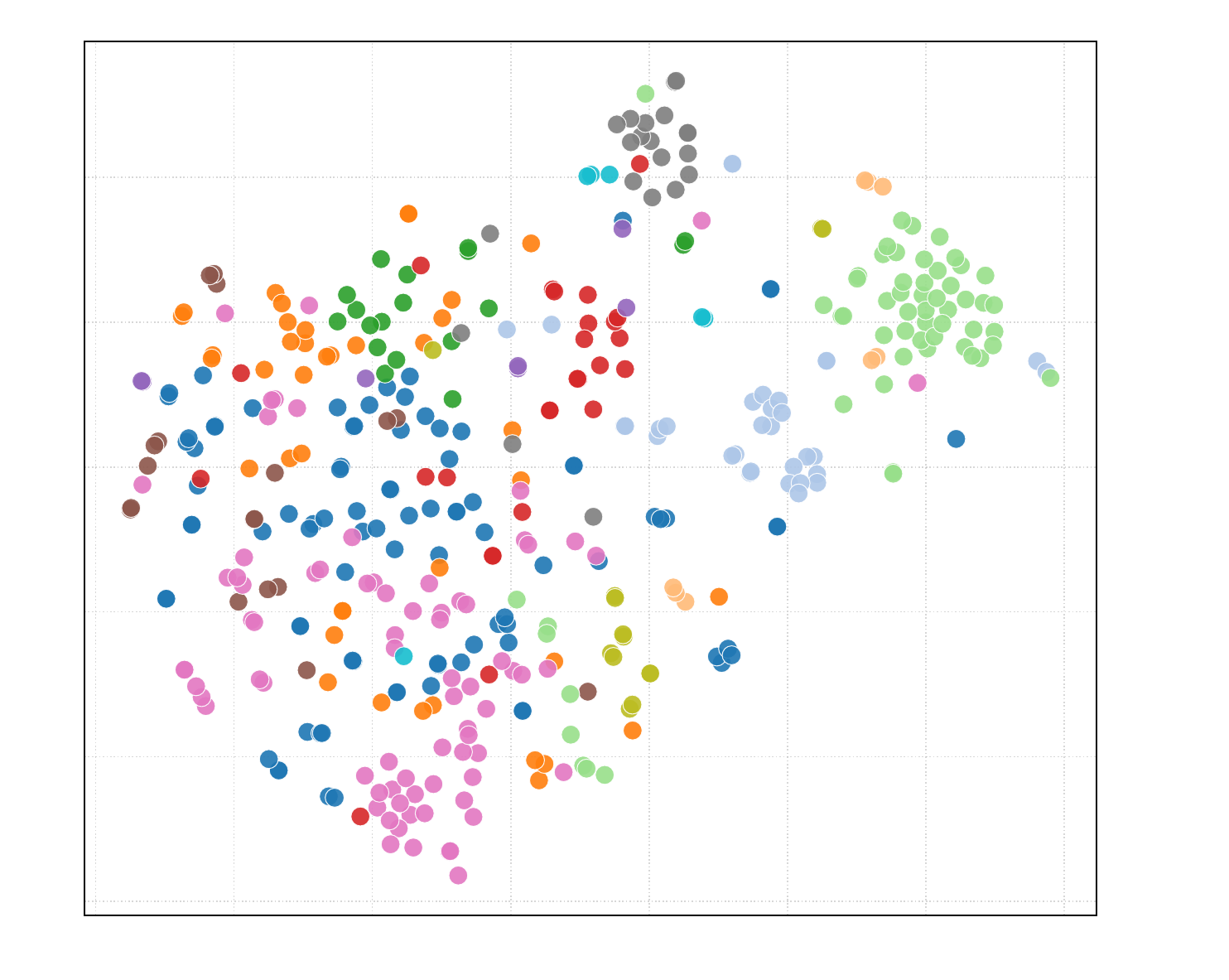} &
\includegraphics[width=0.19\textwidth, trim=48 35 70 20, clip]{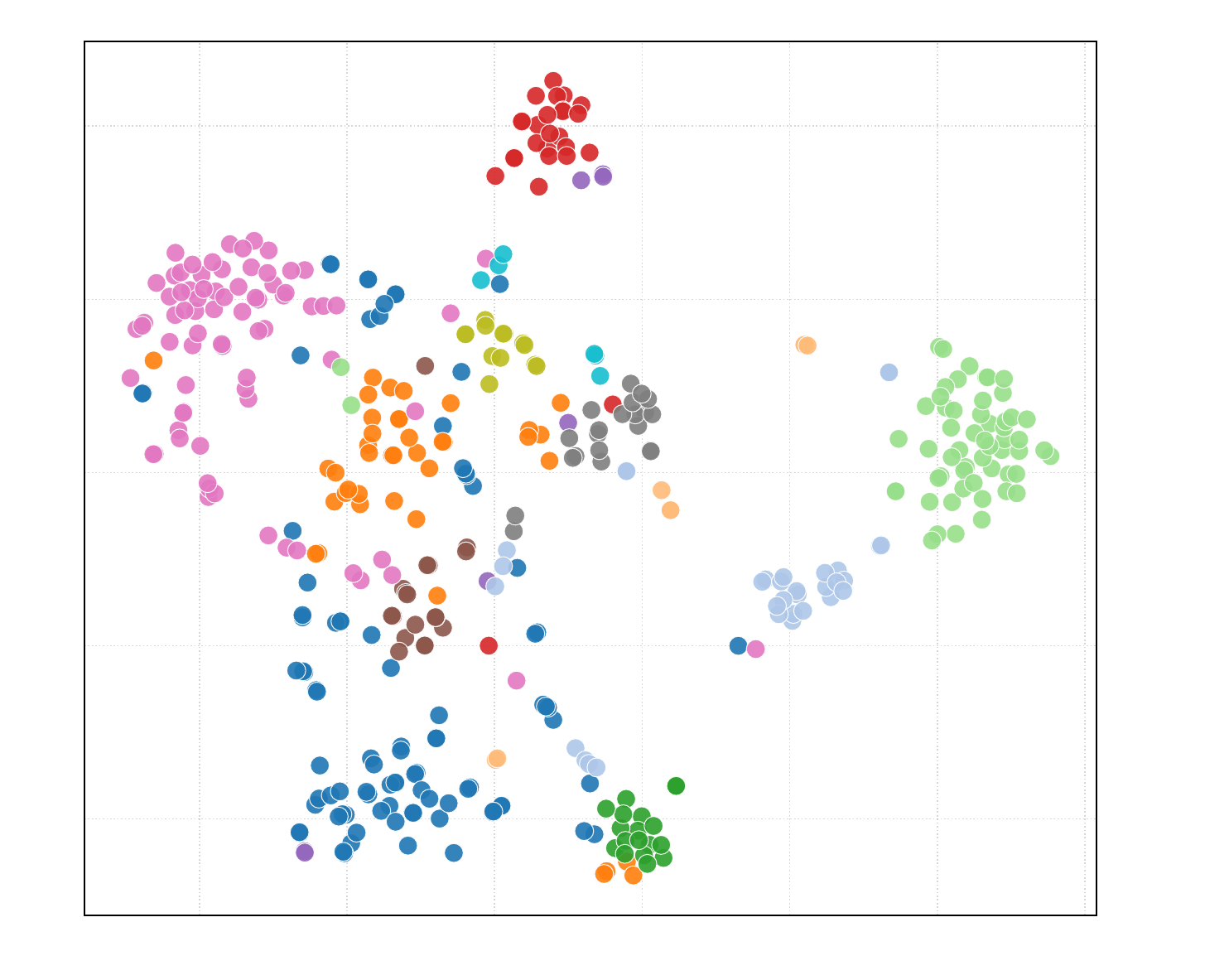} &
\includegraphics[width=0.19\textwidth, trim=48 35 70 20, clip]{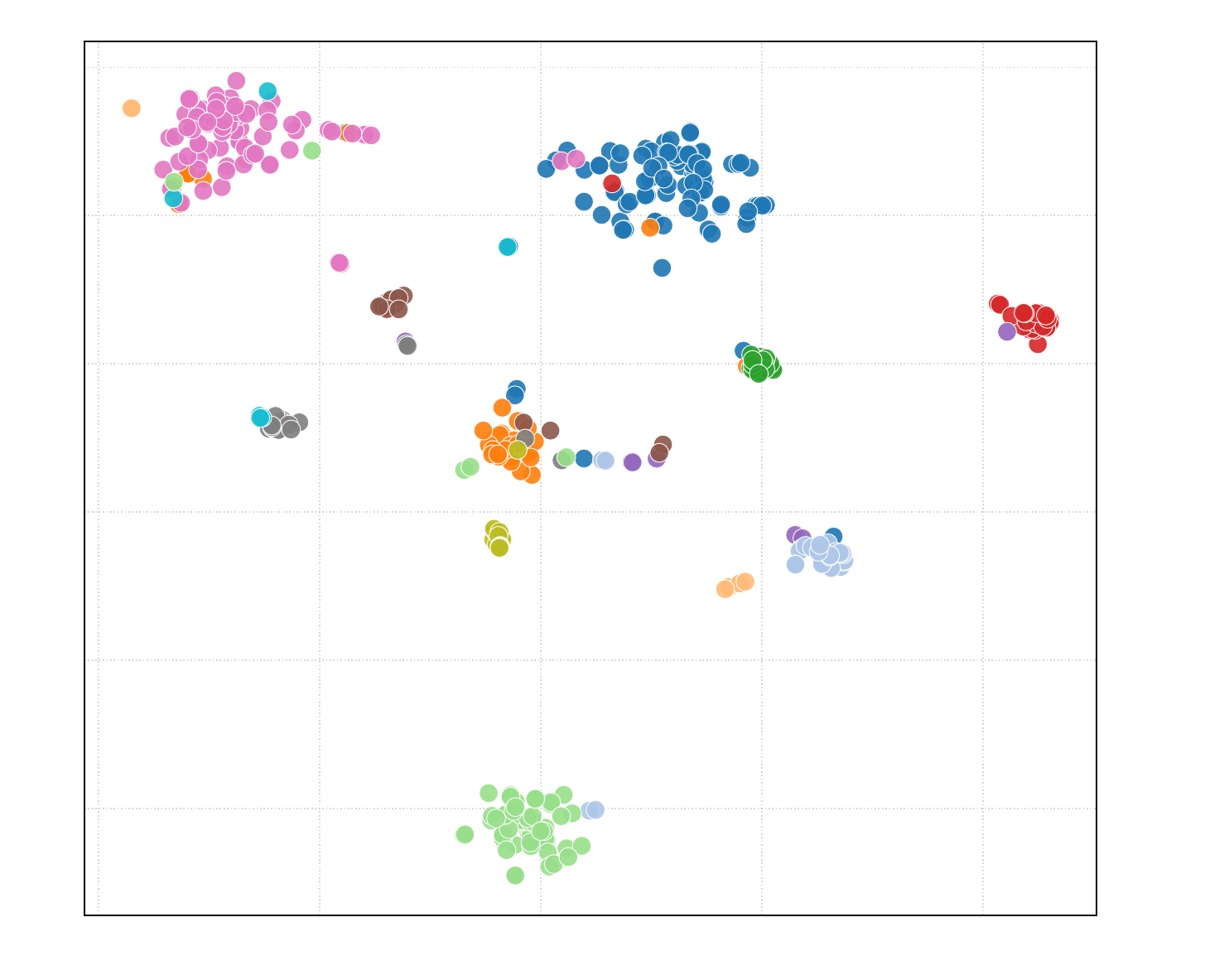}
\end{tabular}
}
\caption{The t-SNE visualization on AS-9K (top) and SLID (bottom) datasets.}
\label{fig:qualitative}
\end{figure*}

\begin{figure*}[t]
\centering
\setlength{\tabcolsep}{2pt}
\resizebox{\textwidth}{!}{
\begin{tabular}{cccccc}
 & ResNet & ViT &  HiFuse & ADSR & Ours \\
\includegraphics[width=0.16\textwidth]{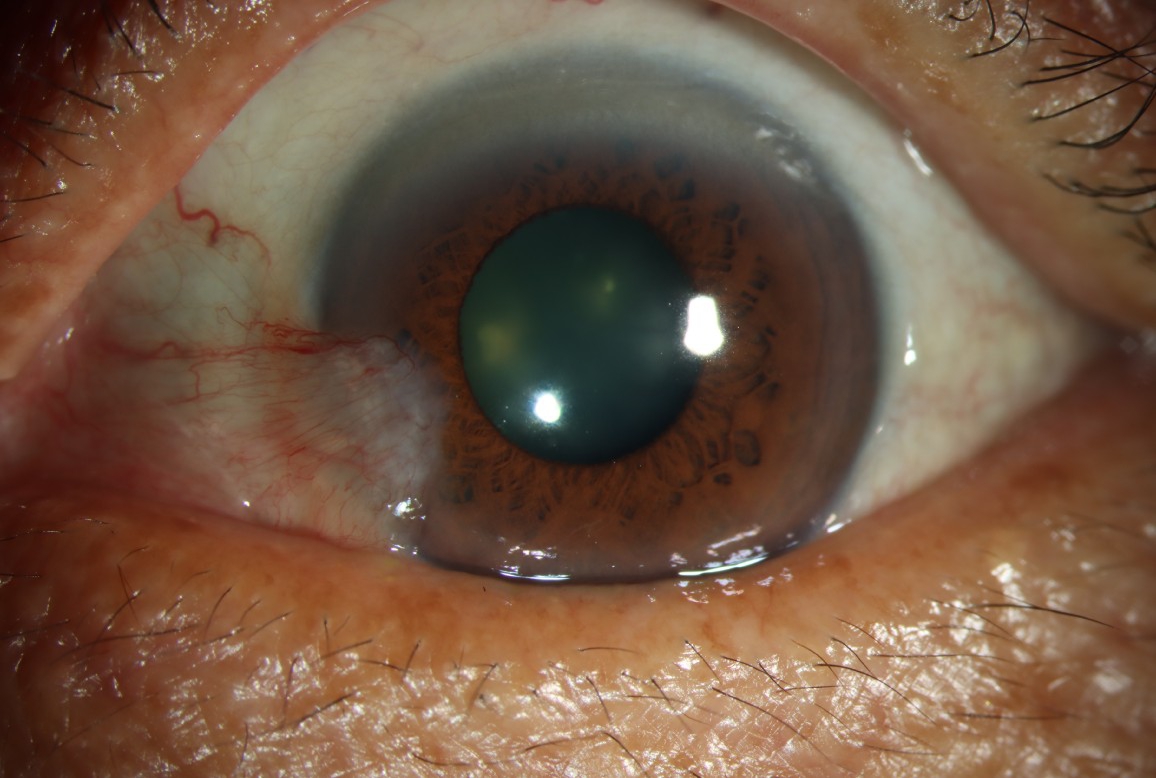} &
\includegraphics[width=0.16\textwidth]{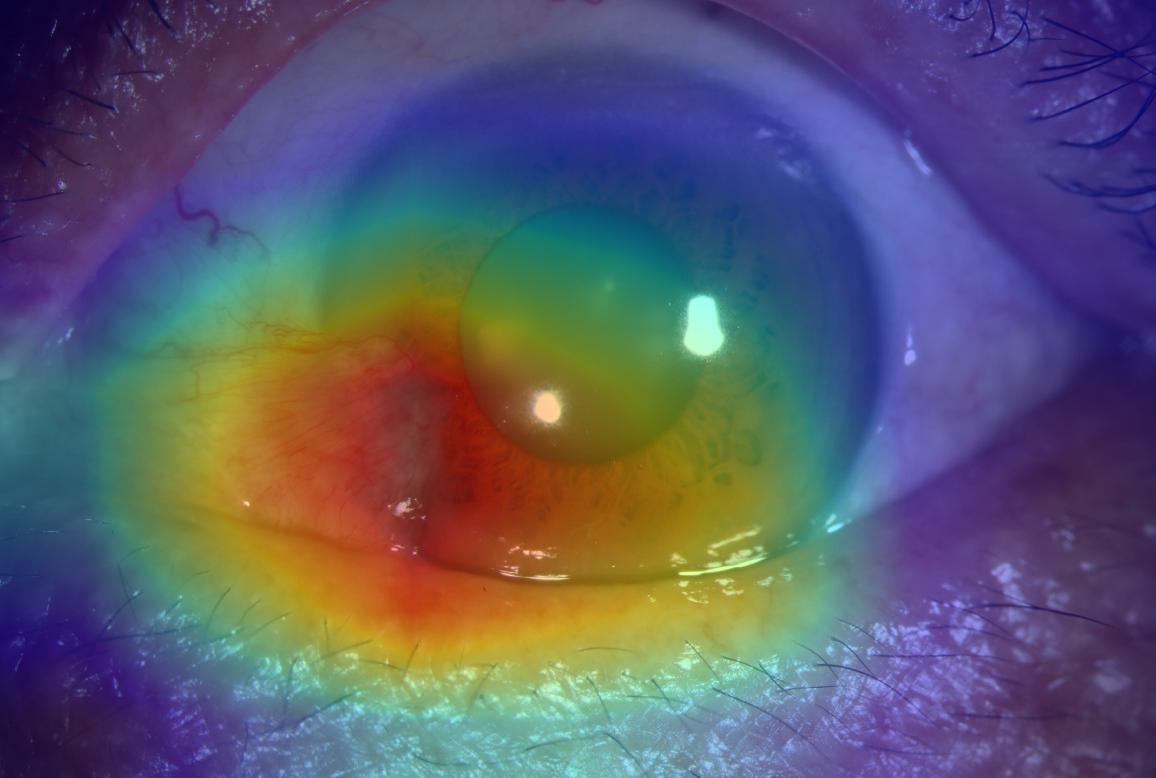} &
\includegraphics[width=0.16\textwidth]{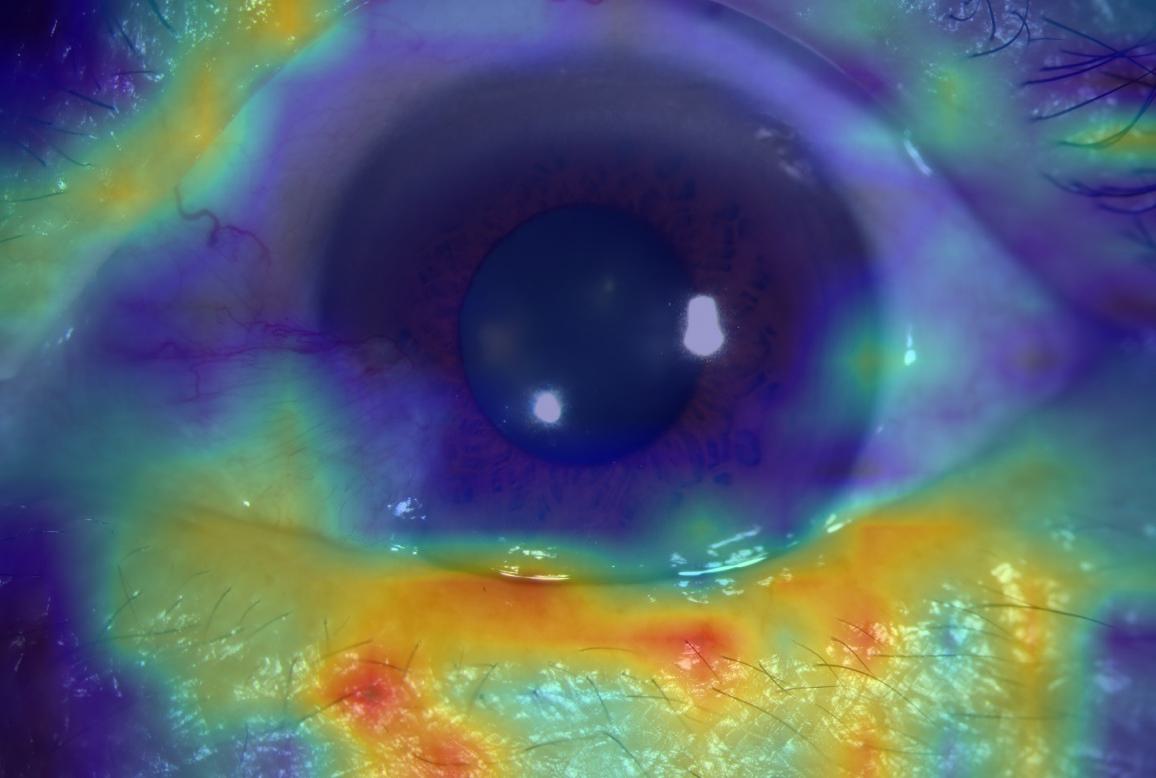} &
\includegraphics[width=0.16\textwidth]{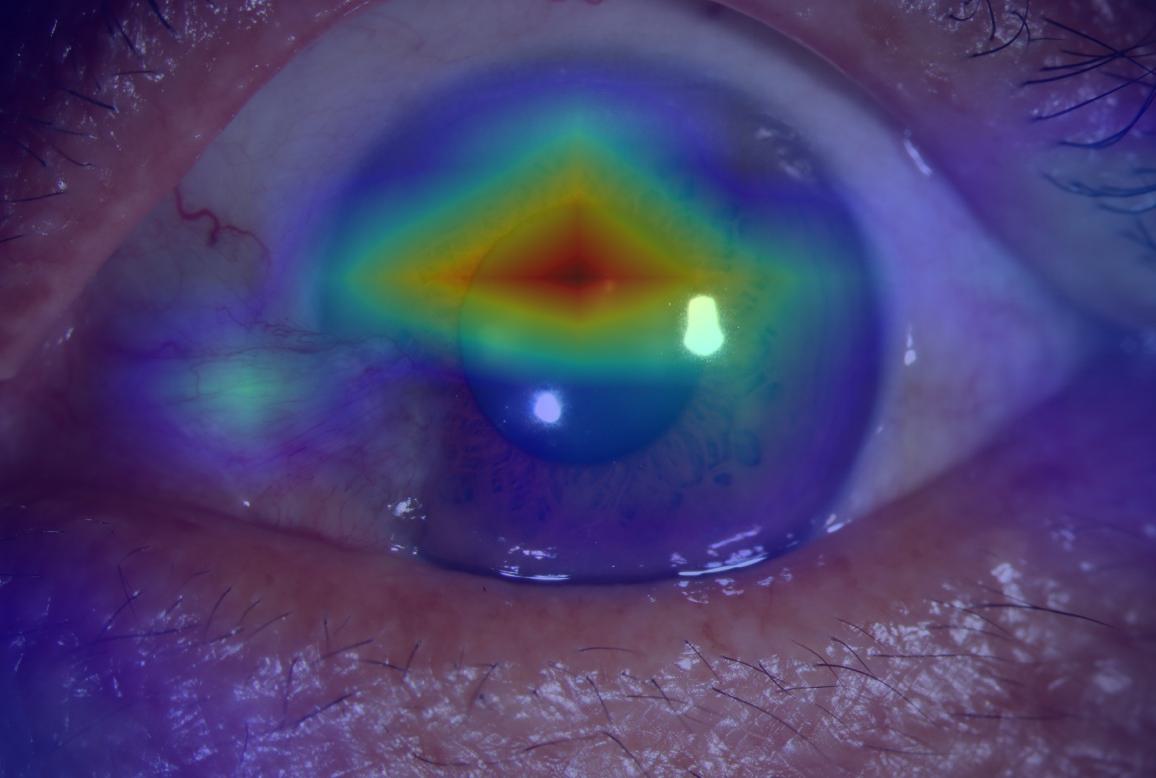} &
\includegraphics[width=0.16\textwidth]{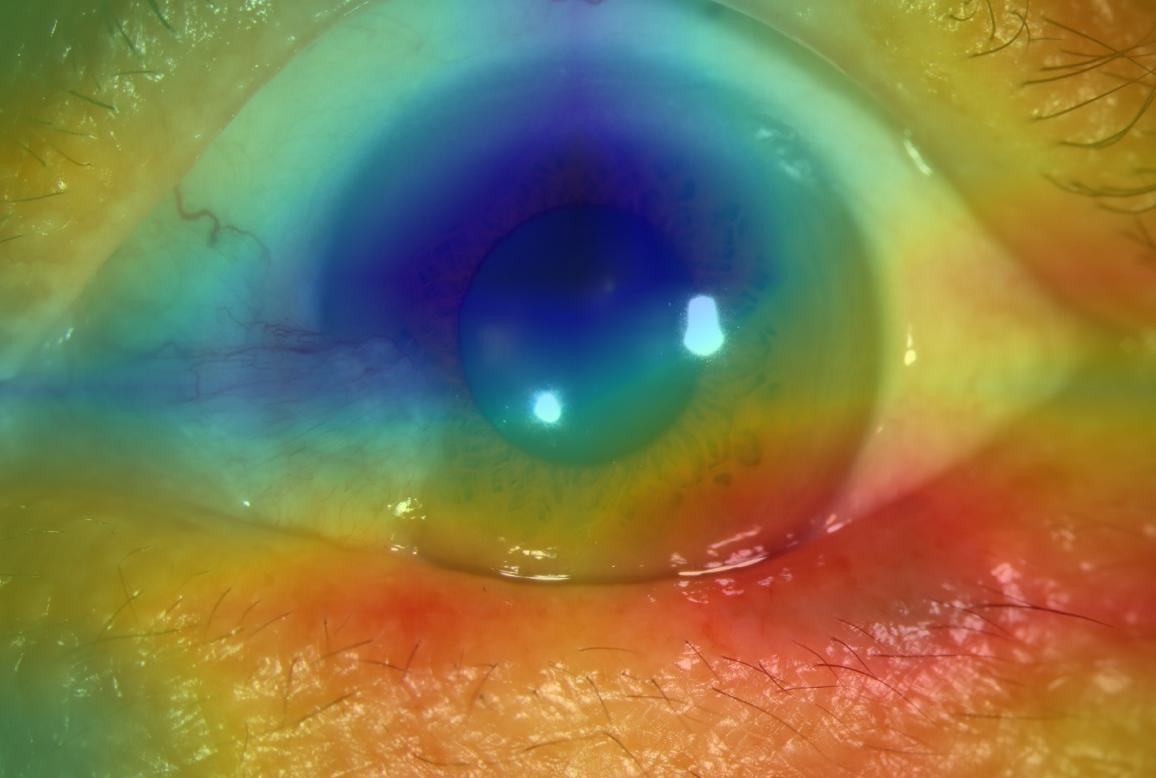} &
\includegraphics[width=0.16\textwidth]{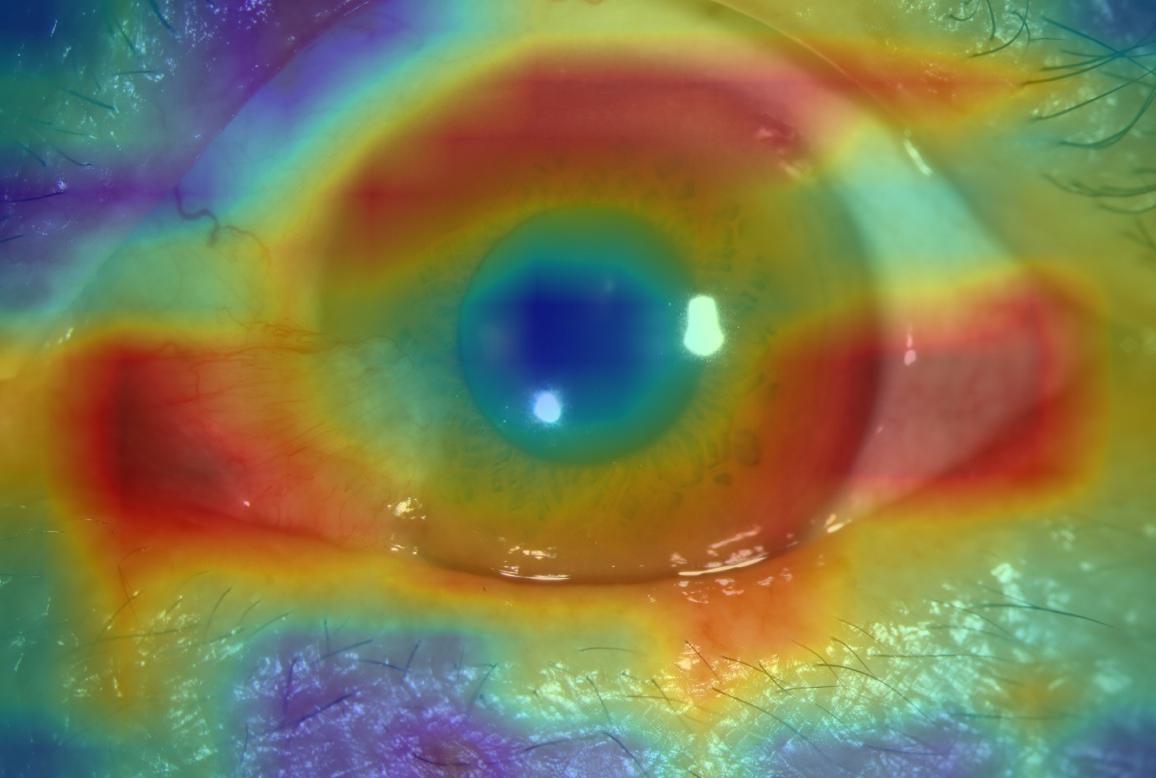} \\

\includegraphics[width=0.16\textwidth]{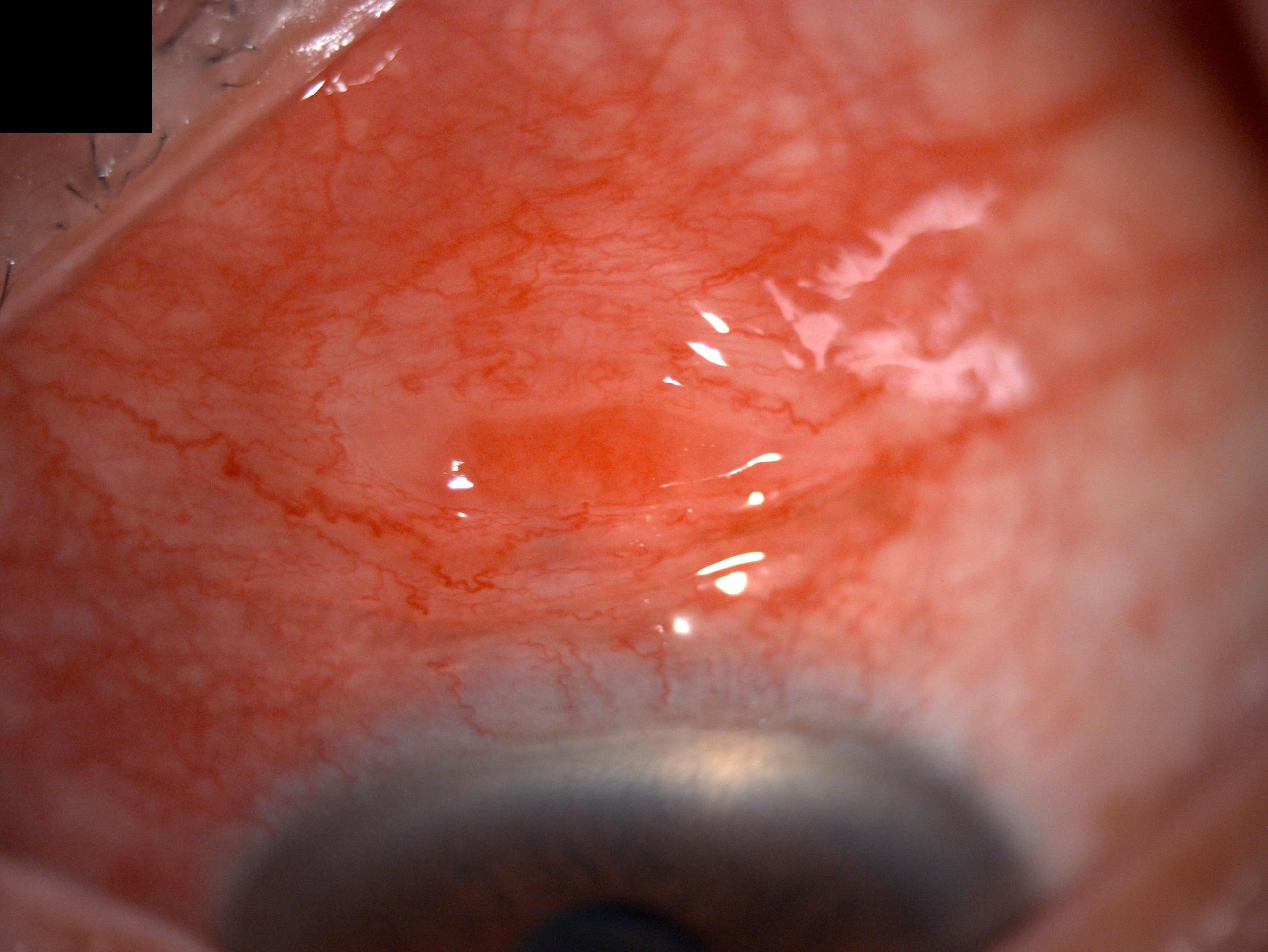} &
\includegraphics[width=0.16\textwidth]{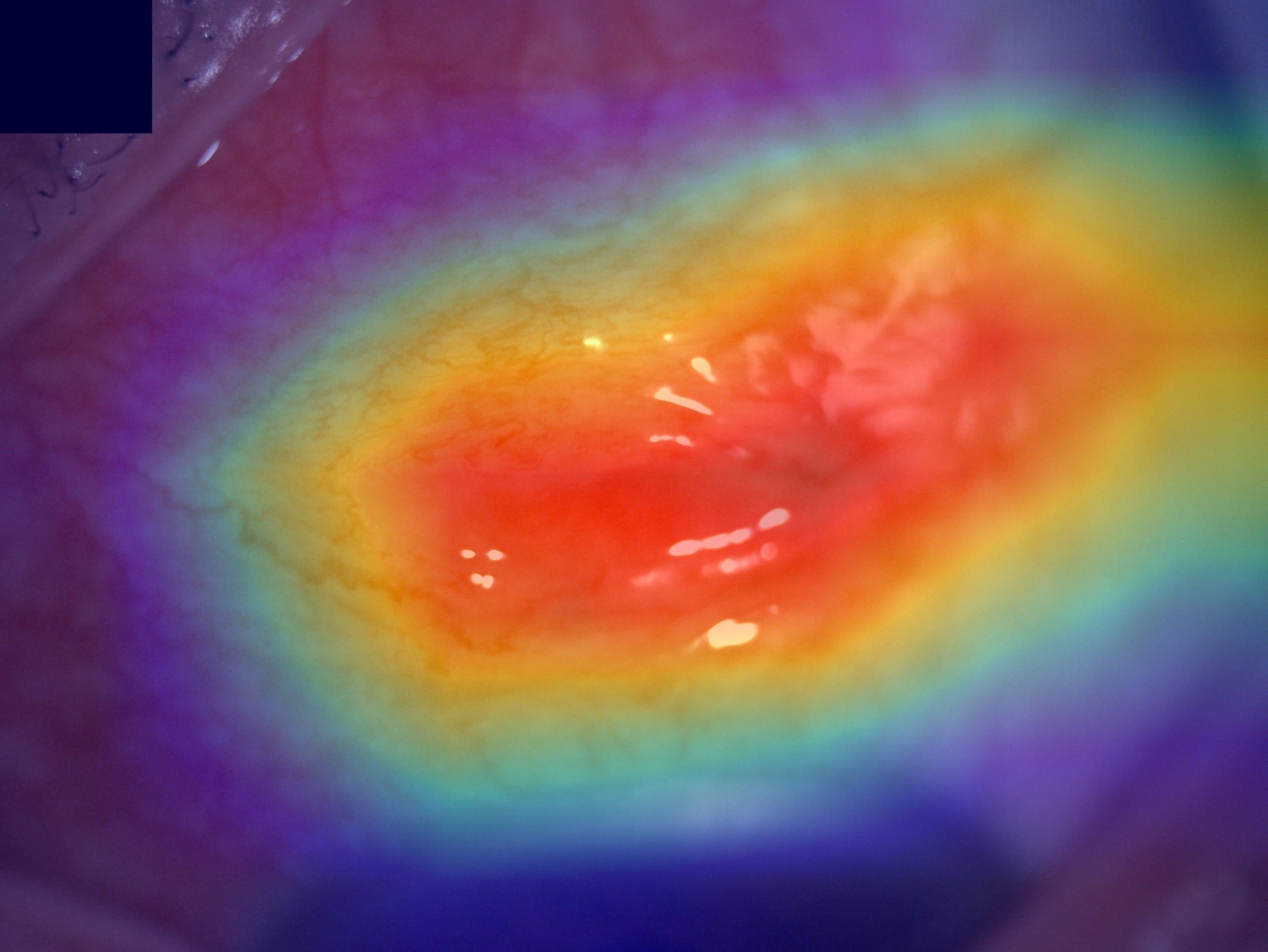} &
\includegraphics[width=0.16\textwidth]{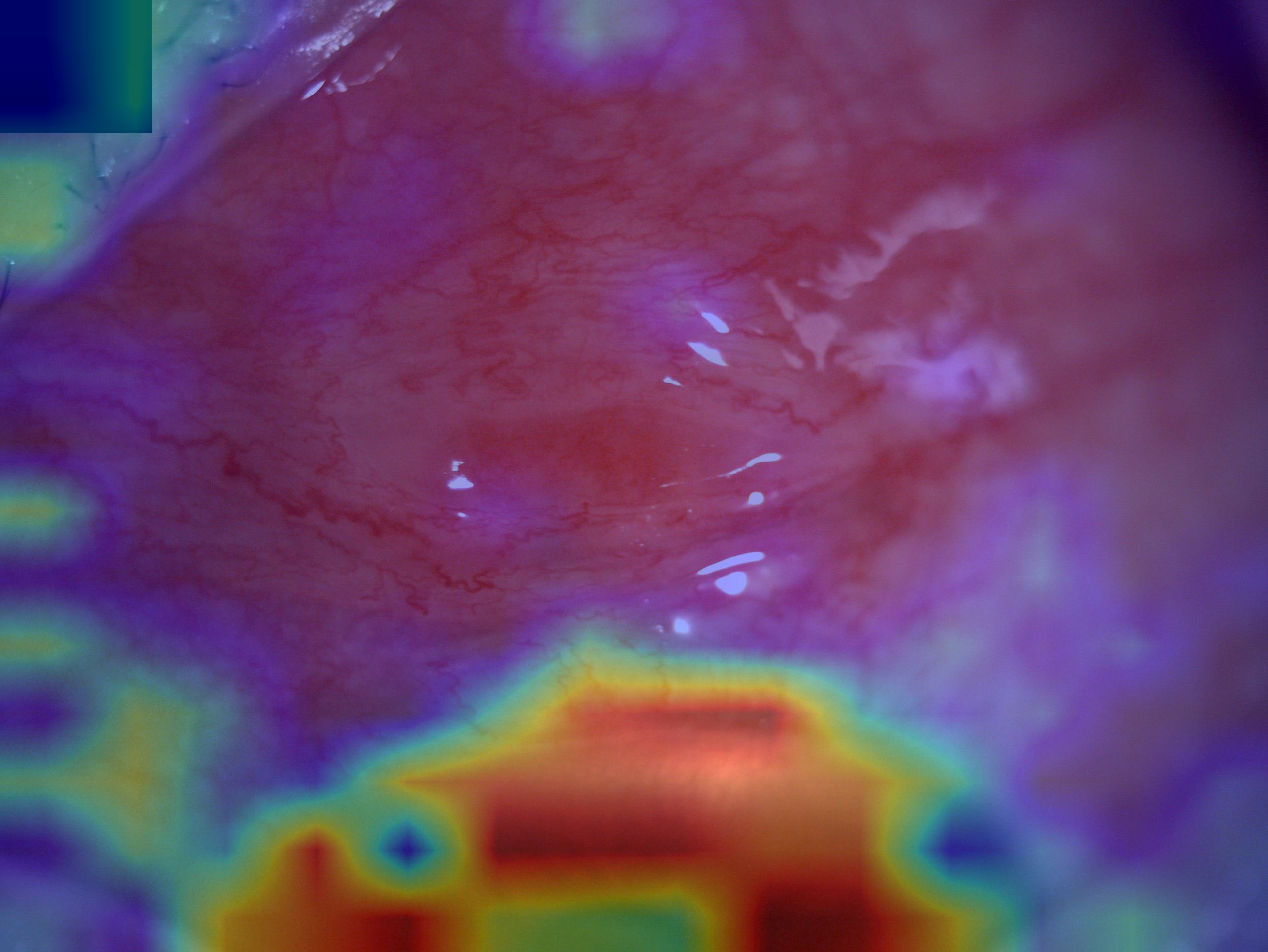} &
\includegraphics[width=0.16\textwidth]{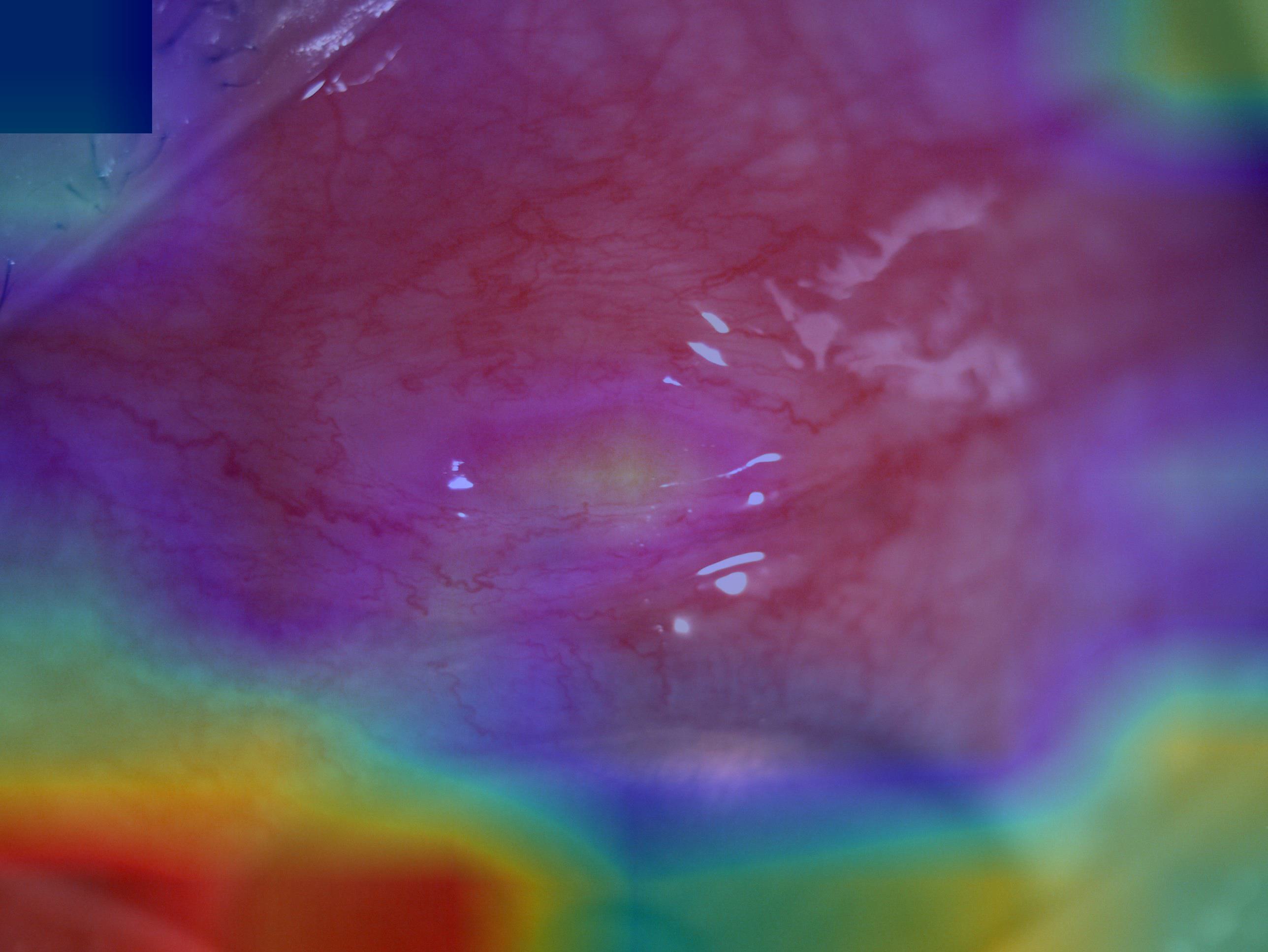} &
\includegraphics[width=0.16\textwidth]{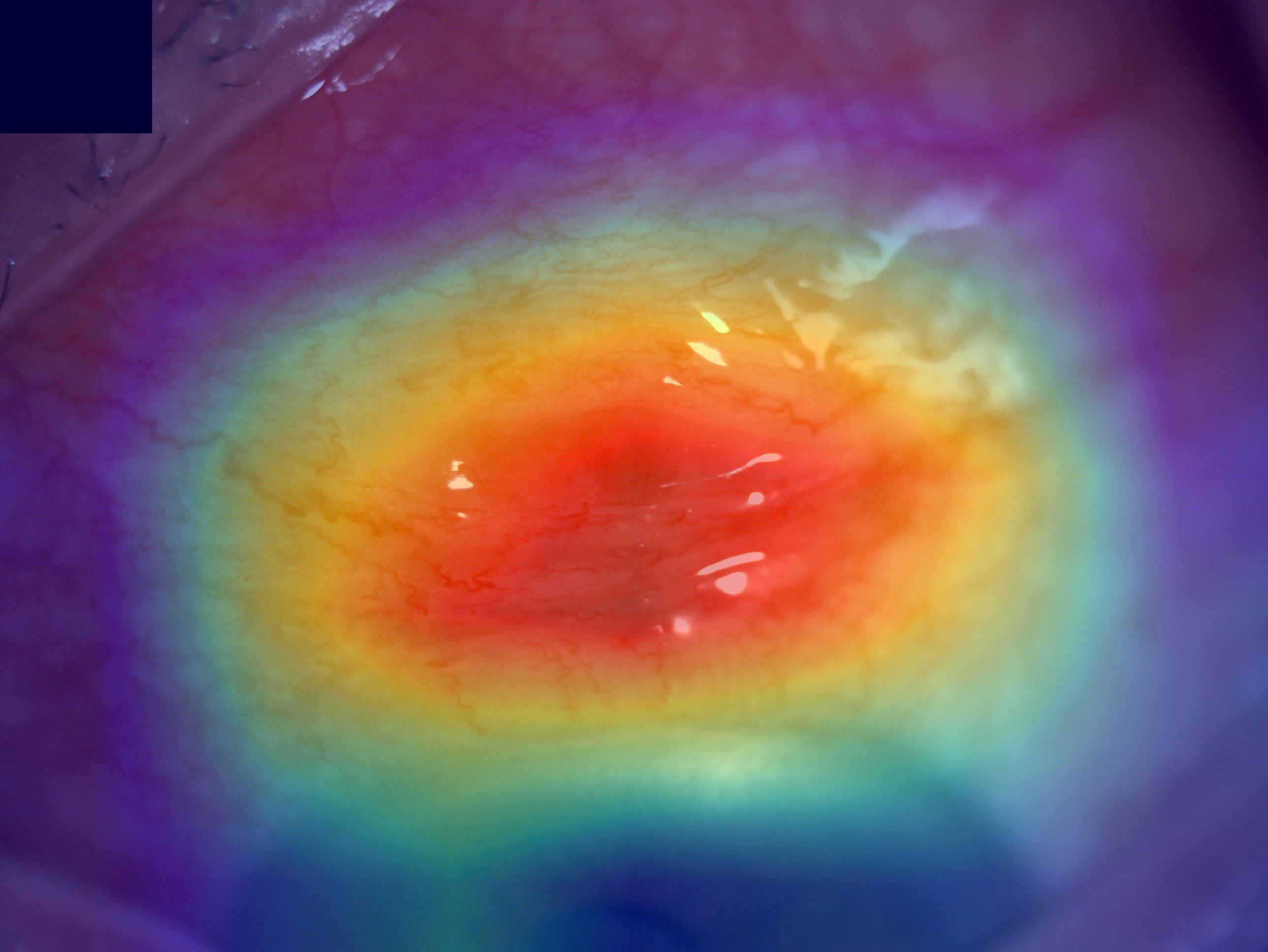} &
\includegraphics[width=0.16\textwidth]{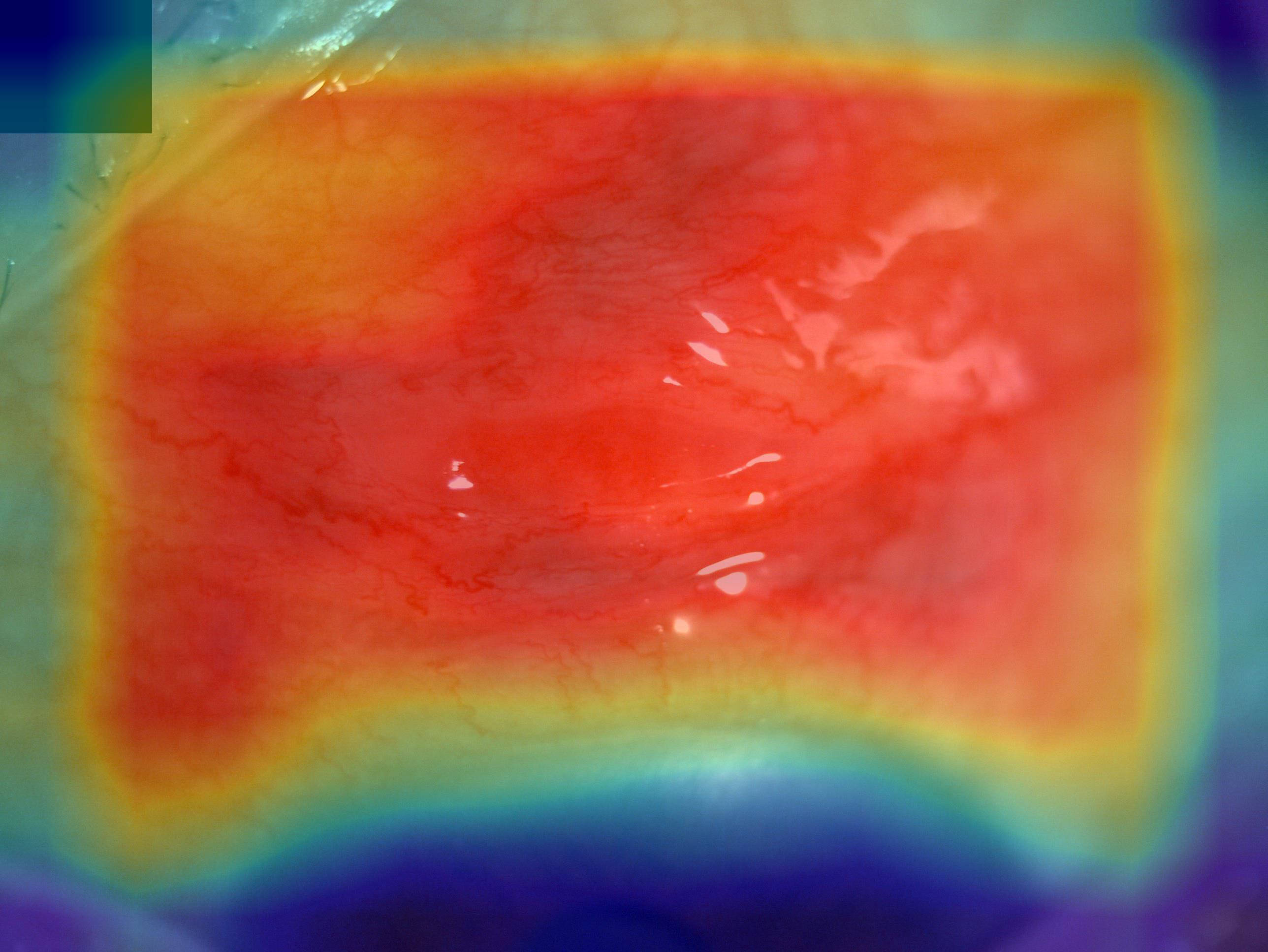}
\end{tabular}
}
\caption{Comparison of Activation Maps on Representative Images from Two Datasets.}
\label{fig:grad}
\end{figure*}


\subsection{Implementation Details}
In the experiments, we adopt ViT-B/16 \cite{dosovitskiy2020image} as the backbone network, initialized with weights pre-trained on ImageNet-21k \cite{russakovsky2015imagenet}. The input images are resized to  224$\times$224.
We use the AdamW \cite{loshchilov2017decoupled} optimizer with an initial learning rate of $10^{-4}$, which is decayed using a cosine annealing schedule. Data augmentation strategies including random horizontal flipping, random rotation, and random erasing are applied during training.
Training is conducted for 100 epochs with a batch size of 64. 
the loss weights are set to $\lambda_c = 0.5$, $\lambda_s = 0.1$, $\lambda_a = 0.1$. 
All experiments are conducted on a single NVIDIA RTX A6000.

\begin{table}[!t]
    \centering
    \renewcommand\arraystretch{1.0}
    \setlength{\tabcolsep}{5pt}
    \caption{Performance comparisons of different methods on AS-9K dataset and SLID dataset. 
    The best and second-best performance metrics are highlighted in bold and underlined, respectively.}
 \resizebox{\columnwidth}{!}{
    \begin{tabular}{lcccccccc}
    \toprule
        \multirow{2}{*}{Method} &
        \multicolumn{4}{c}{\textbf{AS-9K Dataset}} &
        \multicolumn{4}{c}{\textbf{SLID Dataset}} \\
        \cmidrule(r){2-5} \cmidrule(r){6-9}
        & Acc & Precision & Recall & F1
        & Acc & Precision & Recall & F1 \\
    \midrule
        Resnet50 \cite{he2016deep}  
            & 90.10 &81.07 & 76.84 &78.20 
            & 82.48 &77.49 &71.64  &72.08 \\ 

        EfficientNet \cite{efficientnet}   
             & \underline {90.71} &\underline {83.73} & \underline {80.60} &  \underline {80.70}
            & 82.00 & 75.21& \underline {73.33}  & 73.42 \\

        ViT \cite{dosovitskiy2020image}     
              & 87.10 & 71.26 & 72.47& 71.16
             & \underline {84.67} & \underline {80.35} &71.98 & \underline {74.20} \\

        HiFuse \cite{huo2024hifuse}   
             & 88.65 & 79.13 & 75.02 & 76.70
             & 77.37  &77.86 & 60.11 & 64.92\\
         ADSR \cite{zhang2025adaptive}   
             & 89.71 & 81.37& 78.30 &  79.03
             &80.29  &76.82 & 71.65 & 71.64\\
        \rowcolor{gray!20}
        CFCH (ours) 
            & \textbf{91.55} & \textbf{86.53} & \textbf {83.03} &\textbf{83.74}
            & \textbf{86.13} & \textbf{85.90}&\textbf{77.76}  &\textbf{79.58} \\
    \bottomrule
    \end{tabular}}
    \label{tab:classification}
\end{table}

\subsection{Comparisons With SOTA Methods}
To evaluate the effectiveness of our proposed CFCH, we compare our method with recent SOTA methods on two datasets, and conduct both quantitative and qualitative analyses to comprehensively assess its performance. 

\textbf{Quantitative Analysis}
The comparison results are presented in Table \ref{tab:classification}. Our CFCH achieves 83.03\% Recall and 83.74\% F1-score on AS-9K dataset. The Recall and F1-score surpass the second best method by 2.33\% and 3.04\%, respectively.  Our method still outperforms the second best method by 4.43\% and 5.38\% in Recall and F1-score on SLID dataset. These improvements can be attributed to the proposed hierarchical learning framework with a dual-branch parallel design, which effectively exploits anatomical priors and multi-granularity feature alignment. 

\textbf{Qualitative Analysis} To qualitatively evaluate the learned classification representations, feature embeddings are visualized using t-SNE \cite{maaten2008visualizing}. As shown in Fig. \ref{fig:qualitative}, our method produces more compact intra-class clusters and clearer inter-class separation for fine-grained disease categories, indicating improved discriminative capability. These results demonstrate the effectiveness of the proposed hierarchical learning framework in enhancing fine-grained anterior segment disease classification. Moreover, we visualize the model’s attention using Grad-CAM \cite{selvaraju2017grad}, and present visual comparisons on representative images in Fig. \ref{fig:grad}.
Competing methods either fail to fully cover lesion areas or produce diffuse and noisy activation maps. In contrast, CFCH focuses on clinically relevant lesion areas with more precise and compact activations, demonstrating a improved ability to capture fine-grained disease cues and local pathological details.

\begin{table}[!t]
    \centering
    \renewcommand\arraystretch{1.0}
    \setlength{\tabcolsep}{5pt}
    \caption{Ablation study of Multi-Granularity Feature Learning on two datasets. Multi-Granularity represents the combination of coarse-grained and fine-grained branches.}
 \resizebox{\columnwidth}{!}{
    \begin{tabular}{lcccccccc}
    \toprule
        \multirow{2}{*}{Method} &
        \multicolumn{4}{c}{\textbf{AS-9K Dataset}} &
        \multicolumn{4}{c}{\textbf{SLID Dataset}} \\
        \cmidrule(r){2-5} \cmidrule(r){6-9}
        & Acc & Precision & Recall & F1
        & Acc & Precision & Recall & F1 \\
    \midrule
        ViT (Baseline)
             & 87.10 & 71.26 & 72.47& 71.16
             & 84.67 & 80.35&71.98 & 74.20 \\ 

        +Fine-Grained branch  
            & 89.21 &85.24 &75.12  & 77.38
            & 85.64 & 80.50 & 74.56  & 76.62 \\
        +Multi-Granularity  
            & 90.60 & 80.37& 81.62 & 79.46
            & 84.43 &83.31 & 75.98  & 78.07 \\

        \makecell[l]{+Multi-Granularity \\ +Consistency Constraints}
           & \textbf{91.55} & \textbf{86.53} & \textbf {83.03} &\textbf{83.74}
            & \textbf{86.13} & \textbf{85.90}&\textbf{77.76}  &\textbf{79.58} \\ 
    \bottomrule
    \end{tabular}}
    \label{tab:t3}
\end{table}

\begin{table*}[!t]
    \centering
    \renewcommand\arraystretch{1.0}
    \setlength{\tabcolsep}{5pt}
    \caption{Ablation study of different consistency constraints on two datasets.}
    \begin{tabular}{
    >{\centering\arraybackslash}p{20pt}
    >{\centering\arraybackslash}p{20pt}
    >{\centering\arraybackslash}p{32pt}
    >{\centering\arraybackslash}p{32pt}
    c c c c
    c c c c
    }
    \toprule
    \multicolumn{4}{c}{\textbf{Consistency Constraints}} 
    & \multicolumn{4}{c}{\textbf{AS-9K Dataset}} 
    & \multicolumn{4}{c}{\textbf{SLID Dataset}} \\
    \cmidrule(r){1-4} \cmidrule(r){5-8} \cmidrule(r){9-12}
    $\mathcal{L}_{\mathrm{f}}$ 
    & $\mathcal{L}_{\mathrm{c}}$ 
    & $\mathcal{L}_{\mathrm{sem}}$ 
    & $\mathcal{L}_{\mathrm{attn}}$
    & Acc & Precision & Recall & F1
    & Acc & Precision & Recall & F1 \\
    \midrule

    \ding{51} & \ding{51} & - & - 
        & 90.60 & 80.37& 81.62 & 79.46
        & 84.43 &83.31 & 75.98  & 78.07 \\

    \ding{51} & \ding{51} & \ding{51} & - 
        & 91.32 &84.06  & \textbf{83.74} &82.47 
        & 84.91 & 82.37 & 75.01 & 76.56\\

    \ding{51} & \ding{51} & - & \ding{51}
        &91.32  & 84.39 & 83.07 &83.48 
        & \textbf{87.10} & 81.59 & 75.43 & 77.14 \\

    \rowcolor{gray!20}
    \ding{51} & \ding{51} & \ding{51} & \ding{51}
       & \textbf{91.55} & \textbf{86.53} & 83.03 &\textbf{83.74}
           
       & 86.13 & \textbf{85.90}&\textbf{77.76}  &\textbf{79.58}\\

    \bottomrule
    \end{tabular}
    \label{tab:ablation-2}
\end{table*}

\subsection{Ablation Studies}
\noindent \textbf{Analysis of Multi-Granularity Modeling.}
To analyze the effectiveness of multi-granularity modeling, ablation experiments are conducted to evaluate the contribution of each component, as shown in Table \ref{tab:t3}. 
When fine-grained branch is added to the baseline on the AS-9K dataset, it improves ACC and Recall by 2.11\% and 2.65\%, respectively. Similar improvements are also observed on the SLID dataset. 
When coarse-grained and fine-grained features are jointly modeled, further improvements are achieved on both datasets, demonstrating that multi-granularity collaboration enables more comprehensive feature representation and more robust classification. Finally, introducing consistency constraints further enhances performance, yielding the best results across both datasets.
\begin{table}[!t]
\centering
\setlength{\tabcolsep}{5pt}
\renewcommand\arraystretch{1.0}
\caption{Hyperparameter sensitivity analysis of the weighting coefficients $\lambda_c$, $\lambda_s$, and $\lambda_a$ for the proposed loss function on the AS-9K dataset. }
\label{tab:HyperparameterSensitivity}

\begin{tabular}{lccccccc}
\toprule
\multirow{2}{*}{$\lambda_c$} &
\multirow{2}{*}{$\lambda_s$} &
\multirow{2}{*}{$\lambda_a$} &
\multicolumn{4}{c}{\textbf{Classification Performance$\uparrow$}}  \\
\cmidrule(lr){4-7} 
 &  &  & \textbf{ACC} & \textbf{Precision} & \textbf{Recall} & \textbf{F1-score}  \\
\midrule
1   & 1   & 1 & 90.88 & 81.27 & 77.10 & 77.80  \\
0.5   & 1 & 1 & 90.77 & 84.29 & 80.53 & 81.96 \\
0.5 & 0.1 & 0.1 & \textbf{91.55} & \textbf{86.53} & \textbf{83.03} & \textbf{83.74}  \\
0.1   & 1   & 1 & 90.82 & 81.80 & 80.23 & 80.37  \\
0.1 & 0.1   & 0.1 & 90.16 & 84.90 & 78.70 & 79.27  \\

\bottomrule
\end{tabular}
\end{table}

\noindent \textbf{Analysis of Consistency Constraints.}
To investigate the role of consistency constraints in hierarchical learning, we conduct ablation studies on different combinations of loss functions. The baseline setting employs only the fine-grained and coarse-grained classification losses. We then progressively introduce semantic consistency loss $\mathcal{L}_{\mathrm{sem}}$  and cross-granularity attention consistency loss $\mathcal{L}_{\mathrm{attn}}$. As shown in Table \ref {tab:ablation-2}, combining both $\mathcal{L}_{\mathrm{sem}}$ and $\mathcal{L}_{\mathrm{attn}}$ leads to a modest performance improvement, suggesting a partial complementarity between these constraints. Semantic consistency encourages cross-granularity feature alignment, while attention consistency promotes spatial coherence between branches.

\noindent\textbf{Hyperparameter Sensitivity Analysis.}
We further investigate the sensitivity of the proposed framework with respect to the weighting parameters $\lambda_{c}$, $\lambda_{s}$, and $\lambda_{a}$ in the total objective function. As shown in Table \ref{tab:HyperparameterSensitivity}, the proposed model achieves the best performance when $\lambda_{c}=0.5$, $\lambda_{s}=0.1$, and $\lambda_{a}=0.1$, yielding an ACC of 91.55\% and an F1-score of 83.74\%. We observe that both excessively large and excessively small values of the coarse-grained loss weight $\lambda_{c}$ degrade performance, indicating that an appropriate balance between coarse-grained supervision and other optimization objectives is essential. Furthermore, the ablation results in Table \ref{tab:ablation-2} demonstrate that both semantic consistency and attention consistency contribute substantially to learning more discriminative representations. However, assigning overly large weights to these consistency constraints may over-regularize the optimization process and consequently impair model performance. Based on these observations, we adopt $\lambda_{c}=0.5$, $\lambda_{s}=0.1$, and $\lambda_{a}=0.1$ as the default hyperparameter settings in all subsequent experiments.



\section{Conclusion}
In this work, we proposed CFCH, a coarse-fine collaborative hierarchical learning framework for anterior segment image classification. By jointly modeling coarse anatomical context and fine-grained disease semantics with hierarchical consistency learning, CFCH effectively captures structured dependencies across different semantic levels and learns more discriminative representations. We also introduced AS-9K, the largest publicly available dataset for anterior segment image classification, providing a valuable benchmark for future research. Extensive experiments on two datasets demonstrate that CFCH achieves superior performance over existing methods, and visualization results further validate its ability to localize clinically relevant lesions. 
\section*{Acknowledgment}
This work was supported by the Nankai University Eye Institute (No. NKYKK202201), the Scientific Research Project of the Jilin Provincial Department of Education (No. JJKH20230622KJ), the National Natural Science Foundation of China (No. 62272248).

\bibliographystyle{IEEEtran}
\bibliography{IEEEexample}

@article{cai2021eyehealer,
  title={EyeHealer: a large-scale anterior eye segment dataset with eye structure and lesion annotations},
  author={Cai, Wenjia and Xu, Jie and Wang, Ke and Liu, Xiaohong and Xu, Wenqin and Cai, Huimin and Gao, Yuanxu and Su, Yuandong and Zhang, Meixia and Zhu, Jie and others},
  journal={Precision Clinical Medicine},
  volume={4},
  number={2},
  pages={85--92},
  year={2021},
  publisher={Oxford University Press}
}

@inproceedings{wu2024mm,
  title={MM-retinal: Knowledge-enhanced foundational pretraining with fundus image-text expertise},
  author={Wu, Ruiqi and Zhang, Chenran and Zhang, Jianle and Zhou, Yi and Zhou, Tao and Fu, Huazhu},
  booktitle={International Conference on Medical Image Computing and Computer-Assisted Intervention},
  pages={722--732},
  year={2024},
  organization={Springer}
}

@article{zhou2023foundation,
  title={A foundation model for generalizable disease detection from retinal images},
  author={Zhou, Yukun and Chia, Mark A and Wagner, Siegfried K and Ayhan, Murat S and Williamson, Dominic J and Struyven, Robbert R and Liu, Timing and Xu, Moucheng and Lozano, Mateo G and Woodward-Court, Peter and others},
  journal={Nature},
  volume={622},
  number={7981},
  pages={156--163},
  year={2023},
  publisher={Nature Publishing Group UK London}
}

@article{silva2025foundation,
  title={A foundation language-image model of the retina (flair): Encoding expert knowledge in text supervision},
  author={Silva-Rodriguez, Julio and Chakor, Hadi and Kobbi, Riadh and Dolz, Jose and Ayed, Ismail Ben},
  journal={Medical Image Analysis},
  volume={99},
  pages={103357},
  year={2025},
  publisher={Elsevier}
}

@inproceedings{jang2025revisiting,
  title={Revisiting Masked Image Modeling with Standardized Color Space for Domain Generalized Fundus Photography Classification},
  author={Jang, Eojin and Kang, Myeongkyun and Kim, Soopil and Sagong, Min and Park, Sang Hyun},
  booktitle={International Conference on Medical Image Computing and Computer-Assisted Intervention},
  pages={538--548},
  year={2025},
  organization={Springer}
}

@article{li2021preventing,
  title={Preventing corneal blindness caused by keratitis using artificial intelligence},
  author={Li, Zhongwen and Jiang, Jiewei and Chen, Kuan and Chen, Qianqian and Zheng, Qinxiang and Liu, Xiaotian and Weng, Hongfei and Wu, Shanjun and Chen, Wei},
  journal={Nature communications},
  volume={12},
  number={1},
  pages={3738},
  year={2021},
  publisher={Nature Publishing Group UK London}
}

@article{li2024dual,
  title={Dual-mode imaging system for early detection and monitoring of ocular surface diseases},
  author={Li, Yuxing and Chiu, Pak Wing and Tam, Vincent and Lee, Allie and Lam, Edmund Y},
  journal={IEEE Transactions on Biomedical Circuits and Systems},
  volume={18},
  number={4},
  pages={783--798},
  year={2024},
  publisher={IEEE}
}

@inproceedings{xu2019fully,
  title={Fully deep learning for slit-lamp photo based nuclear cataract grading},
  author={Xu, Chaoxi and Zhu, Xiangjia and He, Wenwen and Lu, Yi and He, Xixi and Shang, Zongjiang and Wu, Jun and Zhang, Keke and Zhang, Yinglei and Rong, Xianfang and others},
  booktitle={International Conference on Medical Image Computing and Computer-Assisted Intervention},
  pages={513--521},
  year={2019},
  organization={Springer}
}

@article{kandakji2025hierarchical,
  title={Hierarchical Attention for Sparse Volumetric Anomaly Detection in Subclinical Keratoconus},
  author={Kandakji, Lynn and Woof, William and Pontikos, Nikolas},
  journal={arXiv preprint arXiv:2512.03346},
  year={2025}
}

@article{zamania2023pterygium,
  title={Pterygium Classification Using Deep Patch Region-based Anterior Segment Photographed Images},
  author={Zamania, Nurul Syahira Mohamad and Zakia, W Mimi Diyana W and Huddina, Aqilah Baseri and Mutalibb, Haliza Abdul and Hussaina, Aini},
  journal={Jurnal Kejuruteraan},
  volume={35},
  number={4},
  pages={823--830},
  year={2023}
}

@article{zhang2022machine,
  title={Machine learning for cataract classification/grading on ophthalmic imaging modalities: a survey},
  author={Zhang, Xiao-Qing and Hu, Yan and Xiao, Zun-Jie and Fang, Jian-Sheng and Higashita, Risa and Liu, Jiang},
  journal={Machine Intelligence Research},
  volume={19},
  number={3},
  pages={184--208},
  year={2022},
  publisher={Springer}
}

@inproceedings{he2016deep,
  title={Deep residual learning for image recognition},
  author={He, Kaiming and Zhang, Xiangyu and Ren, Shaoqing and Sun, Jian},
  booktitle={Proceedings of the IEEE conference on computer vision and pattern recognition},
  pages={770--778},
  year={2016}
}

@article{xu7slid,
  title={SLID: A slit-lamp image dataset for deep learning-based anterior eye anatomical segmentation and multi-lesion detection},
  author={Xu, Mingyu and Sun, Yiming and Cheng, Huimin and Zhou, Yifan and Maimaiti, Nuliqiman and Chen, Pengjie and Miao, Qi and Xu, Peifang and Ye, Juan},
  journal={Frontiers in Digital Health},
  volume={7},
  pages={1716501},
  publisher={Frontiers},
 year={2026}
}

@inproceedings{efficientnet,
  title={Efficientnet: Rethinking model scaling for convolutional neural networks},
  author={Tan, Mingxing and Le, Quoc},
  booktitle={International conference on machine learning},
  pages={6105--6114},
  year={2019},
  organization={PMLR}
}

@article{dosovitskiy2020image,
  title={An image is worth 16x16 words: Transformers for image recognition at scale},
  author={Dosovitskiy, Alexey},
  journal={arXiv preprint arXiv:2010.11929},
  year={2020}
}

@misc{dai2024slid,
  author = {Dai, Weiwei and Cheng, Yu and Hu, Min and Xiong, Shibo and Liu, Fei and Cheng, Mianzheng},
  title = {{SLID-E: Slit Lamp Image Dataset for Epiphora - A Benchmark Resource for Automated Tear Overflow Analysis}},
  year = {2024},
  publisher = {figshare},
  howpublished = {Dataset},
  doi = {10.6084/m9.figshare.26172919.v2}
}

@article{son2022deep,
  title={Deep learning-based cataract detection and grading from slit-lamp and retro-illumination photographs: Model development and validation study},
  author={Son, Ki Young and Ko, Jongwoo and Kim, Eunseok and Lee, Si Young and Kim, Min-Ji and Han, Jisang and Shin, Eunhae and Chung, Tae-Young and Lim, Dong Hui},
  journal={Ophthalmology Science},
  volume={2},
  number={2},
  pages={100147},
  year={2022},
  publisher={Elsevier}
}

@inproceedings{ding2025asdc,
  title={ASDC-NET: Anterior Segment Disease Classification Network Based on Slit-Lamp Images},
  author={Ding, Yiming and Zhu, Weifang and Shi, Fei and Ma, Ruiqi and Chen, Xinjian},
  booktitle={2025 IEEE 22nd International Symposium on Biomedical Imaging (ISBI)},
  pages={1--4},
  year={2025},
  organization={IEEE}
}

@inproceedings{chen2023automated,
  title={Automated image quality assessment for anterior segment optical coherence tomograph},
  author={Chen, Boyu and Solebo, Ameenat L and Taylor, Paul},
  booktitle={2023 IEEE 20th International Symposium on Biomedical Imaging (ISBI)},
  pages={1--4},
  year={2023},
  organization={IEEE}
}

@article{sun2024oct,
  title={An AS-OCT image dataset for deep learning-enabled segmentation and 3D reconstruction for keratitis},
  author={Sun, Yiming and Maimaiti, Nuliqiman and Xu, Peifang and Jin, Peng and Cai, Jingxuan and Qian, Guiping and Chen, Pengjie and Xu, Mingyu and Jia, Gangyong and Wu, Qing and others},
  journal={Scientific Data},
  volume={11},
  number={1},
  pages={627},
  year={2024},
  publisher={Nature Publishing Group UK London}
}

@article{maaten2008visualizing,
  title={Visualizing data using t-SNE},
  author={Maaten, Laurens van der and Hinton, Geoffrey},
  journal={Journal of machine learning research},
  volume={9},
  number={Nov},
  pages={2579--2605},
  year={2008}
}

@article{huo2024hifuse,
  title={HiFuse: Hierarchical multi-scale feature fusion network for medical image classification},
  author={Huo, Xiangzuo and Sun, Gang and Tian, Shengwei and Wang, Yan and Yu, Long and Long, Jun and Zhang, Wendong and Li, Aolun},
  journal={Biomedical Signal Processing and Control},
  volume={87},
  pages={105534},
  year={2024},
  publisher={Elsevier}
}

@article{zhang2025adaptive,
  title={Adaptive Dual-Axis Style-Based Recalibration Network With Class-Wise Statistics Loss for Imbalanced Medical Image Classification},
  author={Zhang, Xiaoqing and Xiao, Zunjie and Ma, Jingzhe and Wu, Xiao and Zhao, Jilu and Zhang, Shuai and Li, Runzhi and Pan, Yi and Liu, Jiang},
  journal={IEEE Transactions on Image Processing},
  year={2025},
  publisher={IEEE}
}

@inproceedings{selvaraju2017grad,
  title={Grad-cam: Visual explanations from deep networks via gradient-based localization},
  author={Selvaraju, Ramprasaath R and Cogswell, Michael and Das, Abhishek and Vedantam, Ramakrishna and Parikh, Devi and Batra, Dhruv},
  booktitle={Proceedings of the IEEE international conference on computer vision},
  pages={618--626},
  year={2017}
}

@inproceedings{cai2025retsta,
  title={RetSTA: An LLM-Based Approach for Standardizing Clinical Fundus Image Reports},
  author={Cai, Jiushen and Zhang, Weihang and Liu, Hanruo and Wang, Ningli and Li, Huiqi},
  booktitle={International Conference on Medical Image Computing and Computer-Assisted Intervention},
  pages={544--553},
  year={2025},
  organization={Springer}
}

@article{wang2023transformer,
  title={A transformer-based knowledge distillation network for cortical cataract grading},
  author={Wang, Jinhong and Xu, Zhe and Zheng, Wenhao and Ying, Haochao and Chen, Tingting and Liu, Zuozhu and Chen, Danny Z and Yao, Ke and Wu, Jian},
  journal={IEEE transactions on medical imaging},
  volume={43},
  number={3},
  pages={1089--1101},
  year={2023},
  publisher={IEEE}
}

@article{loshchilov2017decoupled,
  title={Decoupled weight decay regularization},
  author={Loshchilov, Ilya and Hutter, Frank},
  journal={arXiv preprint arXiv:1711.05101},
  year={2017}
}

@article{li2021digital,
  title={Digital technology, tele-medicine and artificial intelligence in ophthalmology: A global perspective},
  author={Li, Ji-Peng Olivia and Liu, Hanruo and Ting, Darren SJ and Jeon, Sohee and Chan, RV Paul and Kim, Judy E and Sim, Dawn A and Thomas, Peter BM and Lin, Haotian and Chen, Youxin and others},
  journal={Progress in retinal and eye research},
  volume={82},
  pages={100900},
  year={2021},
  publisher={Elsevier}
}

@article{jin2024hmil,
  title={HMIL: hierarchical multi-instance learning for fine-grained whole slide image classification},
  author={Jin, Cheng and Luo, Luyang and Lin, Huangjing and Hou, Jun and Chen, Hao},
  journal={IEEE Transactions on Medical Imaging},
  volume={44},
  number={4},
  pages={1796--1808},
  year={2024},
  publisher={IEEE}
}

@article{ran2023comprehensive,
  title={Comprehensive survey on hierarchical clustering algorithms and the recent developments},
  author={Ran, Xingcheng and Xi, Yue and Lu, Yonggang and Wang, Xiangwen and Lu, Zhenyu},
  journal={Artificial Intelligence Review},
  volume={56},
  number={8},
  pages={8219--8264},
  year={2023},
  publisher={Springer}
}

@article{russakovsky2015imagenet,
  title={Imagenet large scale visual recognition challenge},
  author={Russakovsky, Olga and Deng, Jia and Su, Hao and Krause, Jonathan and Satheesh, Sanjeev and Ma, Sean and Huang, Zhiheng and Karpathy, Andrej and Khosla, Aditya and Bernstein, Michael and others},
  journal={International journal of computer vision},
  volume={115},
  pages={211--252},
  year={2015},
  publisher={Springer}
}

@article{zhang2026ai,
  title={AI framework for multidisease detection via retinal imaging},
  author={Zhang, Xiayin and Li, Qinyi and Liang, Yinhao and Lai, Chunran and Cao, Jiahui and Feng, Yangqin and Hu, Wenyi and Jiang, Hongyang and Liu, Chunxin and Zhang, Feng and others},
  journal={Nature Medicine},
  pages={1--10},
  year={2026},
  publisher={Nature Publishing Group US New York}
}

\end{document}